\documentclass[10pt,journal,compsoc]{IEEEtran}

\usepackage{subfigure}
\usepackage{graphicx}
\usepackage{epsfig}
\usepackage{amsmath}
\usepackage{cite}

\usepackage{verbatim}
\usepackage{array}
\usepackage{placeins}
\usepackage{algorithm}
\usepackage{algorithmic}
\usepackage{setspace}
\usepackage{balance}
\usepackage{picins}
\usepackage{eqparbox}
\usepackage[export]{adjustbox}

\makeatother

\begin{document}

\author{Jun~Xu,~\IEEEmembership{Student Member,~IEEE}, G\"urkan~Solmaz,~\IEEEmembership{Member,~IEEE}, Rouhollah~Rahmatizadeh,~\IEEEmembership{Student Member,~IEEE}, Damla~Turgut,~\IEEEmembership{Member,~IEEE} and Ladislau~B{\"o}l{\"o}ni,~\IEEEmembership{Senior Member,~IEEE}%

\IEEEcompsocitemizethanks{

\IEEEcompsocthanksitem All the authors are with the Department of Computer Science, University of Central Florida, Orlando,
FL, 32816.\protect\\
E-mail: \{junxu,gsolmaz,rrahmati,turgut,lboloni\}@eecs.ucf.edu\protect
\IEEEcompsocthanksitem Part of this work was published in IEEE LCN 2015 ~\cite{7366291}
}}

\markboth{}%
{Xu \MakeLowercase{\textit{et al.}}: Bare Demo of IEEEtran.cls for Journals}

\title{Internet of Things Applications: Animal Monitoring with Unmanned Aerial Vehicle}
\IEEEtitleabstractindextext{%

\begin{abstract}

In animal monitoring applications, both animal detection and their movement prediction are major tasks. While a variety of animal monitoring strategies exist, most of them rely on mounting devices. However, in real world, it is difficult to find these animals and install mounting devices. In this paper, we propose an animal monitoring application by utilizing wireless sensor networks (WSNs) and unmanned aerial vehicle (UAV). The objective of the application is to detect locations of endangered species in large-scale wildlife areas and monitor movement of animals without any attached devices. In this application, sensors deployed throughout the observation area are responsible for gathering animal information. The UAV flies above the observation area and collects the information from sensors. To achieve the information efficiently, we propose a path planning approach for the UAV based on a Markov decision process (MDP) model. The UAV receives a certain amount of reward from an area if some animals are detected at that location. We solve the MDP using Q-learning such that the UAV prefers going to those areas that animals are detected before. Meanwhile, the UAV explores other areas as well to cover the entire network and detects changes in the animal positions. We first define the mathematical model underlying the animal monitoring problem in terms of the value of information (VoI) and rewards. We propose a network model including clusters of sensor nodes and a single UAV that acts as a mobile sink and visits the clusters. Then, one MDP-based path planning approach is designed to maximize the VoI while reducing message delays. The effectiveness of the proposed approach is evaluated using two real-world movement datasets of zebras and leopard. Simulation results show that our approach outperforms greedy and random heuristics as well as the path planning based on the solution of the traveling salesman problem.

\end{abstract}

\begin{IEEEkeywords}
value of information; markov decision process; path planning; animal monitoring; unmanned aerial vehicles.
\end{IEEEkeywords}}

\maketitle
\IEEEdisplaynontitleabstractindextext
\IEEEpeerreviewmaketitle

\section{Introduction}
\label{sec:Introduction}

Wireless sensor networks (WSNs) have been widely used for wildlife monitoring and tracking of different species~\cite{6297993}. Often, different functional devices working as sensors are used to report animal data to the base station~\cite{5930512}~\cite{Juang02}. However, animal monitoring is difficult in remote large-scale wildlife areas due to the dangerous environment and the uncertainty of animal movement patterns. In addition, due to the energy limitations of traditional sensor networks, sending the sensed information in a timely manner is costly and impractical.

The recent advances in technology of unmanned aerial vehicles (UAVs) allow their usage as a part of the WSNs. UAVs provide an extremely flexible platform for WSNs applications by playing different roles in WSNs such as actors~\cite{6297993}, sensors~\cite{4301324}, and mobile sinks~\cite{akbas2012actor}. In some monitoring scenarios such as forest fire monitoring or animal monitoring, the sensed information is time-sensitive. In other words, the earlier the sensed information is reported to a base station, the higher value the information has. The arise of UAVs provide cost-effective and appealing solutions for surveillance applications.



In this paper, we focus on the animal monitoring and tracking in large wild areas. In an unknown large area, it is difficult and sometimes infeasible to find wildlife animals and attach wearable tracking devices to them. In this case, low cost sensors can be widely deployed throughout the observation area to detect and recognize the wildlife appearance. Currently, sensors can identify animals appearance by different types of inputs such as smell, sound and image. With deployed sensors, we can get animal appearance information in wildlife areas, however, long distance wireless communication in large remote wildlife areas is costly and impractical. If the delay of reporting sensed information becomes too large, the animal may leave that place. Therefore, the key challenge is to collect and transmit the sensed animal data in a timely and efficient way such that we can track and monitor those animals. By observing animal activities, we find that animal activities usually have some specific features such as living in groups and having more activities around habitats. These features make animal movements have some specific patterns. Examples of animal movement trajectories can be seen in ~\ref{sec:sec:Network Model}, instead of having activities in a large area, animals usually have activities only in one or several small areas. If we focus more on these "hot" areas, we have a much higher probability to track and predict animals movement. Inspired by this idea, we divide the whole observation area into small virtual grids. Then based on this grid structure network, we design our system model by utilizing a UAV and letting the UAV to explore and learn these "hot" small virtual grids. By treating each grid as a cluster of sensor nodes, data collection by the UAV is reduced to visiting the cluster head of each grid. The Markov decision process (MDP) model is used to do path planning for the UAV as each grid represents a state of MDP. We solve the MDP model using Q-learning algorithm by letting UAV to receive rewards from grids when animal data is reported. The goal of our path planning approach is collecting animal data efficiently and then predicting future animal movements. The UAV is self-learning and can dynamically plan and adapt according to environment changes.

To quantitatively weigh the value of the sensed animal appearance information, the proposed path planning approach utilizes the value of information (VoI)~\cite{turgut2013ive}. The basic idea of VoI is that the sensed information has the highest valued when it is first generated and the value decrease as time goes. In such a way, the task of UAV can be reduced to maximizing the overall value of information obtained from the entire network. We give a mathematical model for calculating the VoI in animal monitoring operation in Section~\ref{sec:Mathematical Model}. Lastly, we evaluate the performance of the proposed network model and the path planning approach with real animal datasets: ZebraNet~\cite{Juang02} and leopard tortoises dataset.




The remainder of this paper is organized as follows. Section~\ref{sec:Mathematical Model} defines mathematical model for calculating the value of information as well as the reward in Q-learning. Section~\ref{sec:Network Model} defines the network model and explains the implementation of Markov decision process (MDP) for UAV path planning. Section~\ref{sec:Simulation Study} evaluates the performance of the proposed network model and the path planning approach. Section~\ref{sec:RelatedWorks} briefly summarizes the related studies on animal tracking and mobile sinks path planning in WSNs. Finally, Section~\ref{sec:Conclusion} gives the conclusion.

\section{Mathematical Model}
\label{sec:Mathematical Model}

In this section, we describe the mathematical model of the value of information for animal monitoring and our definition of the sensors' credibility and initial rewards.

\subsection{Value of information}
{\em Value of information} (VoI) is a metric initially proposed in game theory as the price an optimal player would pay for a piece of information. This metric is used in recent studies of sensor networks~\cite{turgut2013ive} as a way of assigning a higher value to more recently sensed data. The VoI of an event is highest at the moment the event is created and continuously decreases as time passes.

Let us describe an example environmental mission where VoI has critical importance. In the times of marine oil spills, where crude oil is released into ocean or coastal waters from offshore platforms or tankers, actions must be taken to stop leakage and repair the damages at the earliest time. After the oil leakage information is sensed by a set of underwater sensor nodes, early arrival of the information to the base station will provide the operators more ability to take decisions for repairing and patching up the leaks.

To the best of our knowledge, the concept of VoI is not considered for the animal monitoring problem in the literature. However, maximizing VoI may be very useful for monitoring wild animals such as finding the current locations of endangered species. For instance, earlier arrival of the information to a UAV that collects data may be helpful for finding the exact location of an animal who needs certain help or rescue from a region. Moreover, earlier information retrieval by the UAV may result in direct observation of the animal.

The idea behind VoI can be described by a scenario where an actuation action has to be taken on the basis of sensed data. The sensed information is of more value at present as compared to it being processed for actuation at a later time. For many scenarios, we define the value of information in terms of an exponential decay (although other forms are possible):\par

\begin{equation}
\label{Eq:VoI}
F_{VoI}(t)=Ae^{-Bt}
\end{equation}

\begin{figure}[htbp]
\centering\includegraphics[width=0.45\textwidth]{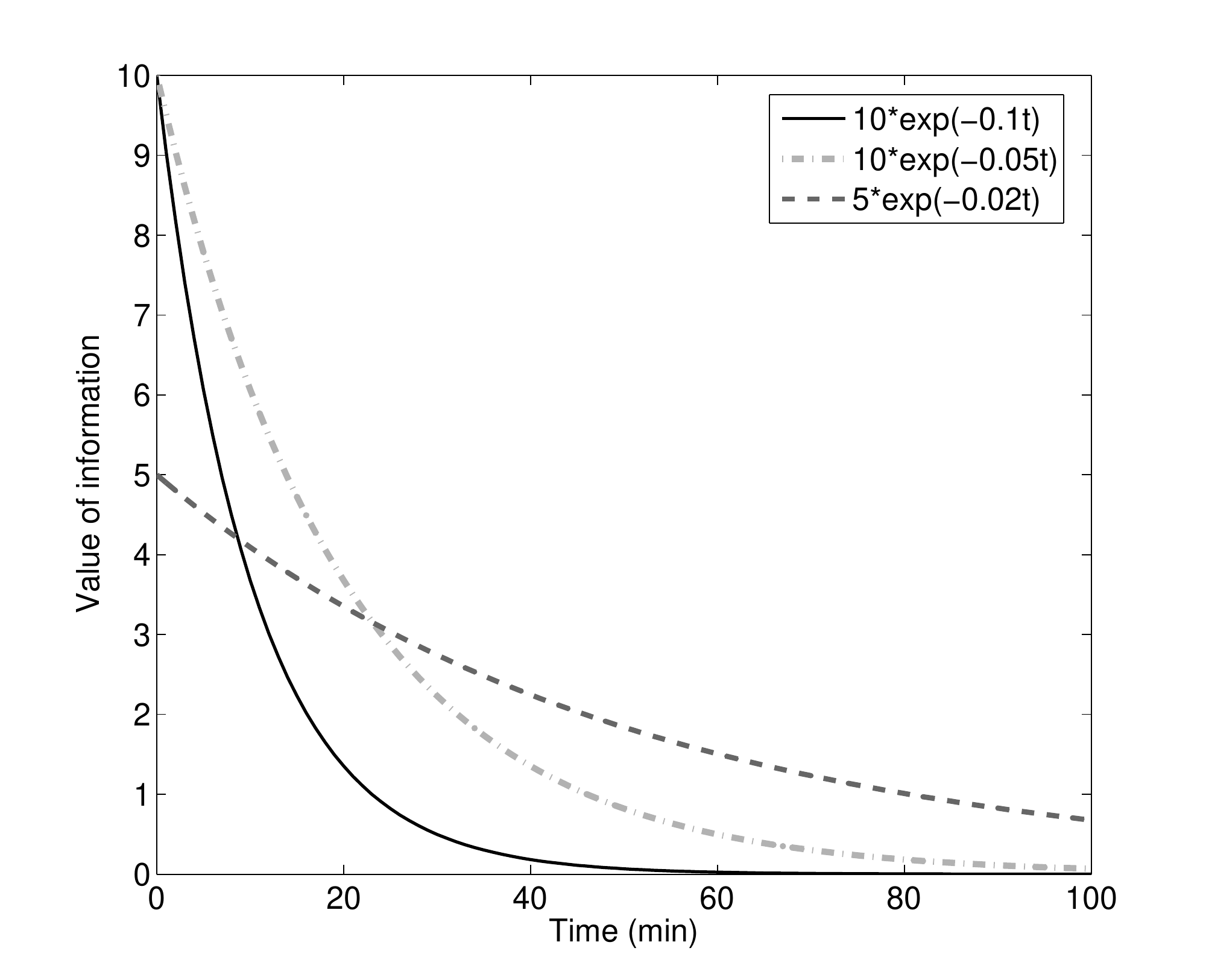}
\caption{VoI examples for different A and B values.}\label{fig:1}
\end{figure}

In Equation~\ref{Eq:VoI}, the constant value A represents the initial value of the information while B represents the decay speed of the VoI. A higher value of A defines the information with a higher initial value while a higher value of B defines a faster decay of the VoI. Fig.~\ref{fig:1} shows three different examples of A and B values and their outcomes as the VoI function. It may be considered as representations of three urgent levels of data in a sensor network. As it can be seen in the figure, higher values such as $A=10$ and $B=0.1$ produce the VoI with sharper decays, meaning urgent events that lose their importance earlier. If we define a threshold value of events as $F_{VoI}=1$, the events with higher values of A and B expire before 60 minutes while the events with lower values ($A=5$, $B=0.02$) expire around 90 minutes.

\subsection{Initial rewards}

We define each sensed information as an {\em event} in the system. Each event has an {\em initial reward} (IR) that is the value of information at the time of the event occurs ($t=0$). In other words, IR is the maximum value of information that can be gathered for each detected event. In our application, IR is affected by several factors: credibility of sensors, distance to the animal and the duration.


{\em Credibility} (C) is a parameter that represents the reliability of the sensed information. We define credibility based on heterogeneous sensor nodes with different abilities and quality of services. For instance, an acoustic sensor node detects the sounds coming from an animal while another sensor node can take visual images. Moreover, a sensor node may have a lower quality camera that results in lower resolution images than the other sensor nodes. We assume there are $k$ sensor nodes $\{N_1, N_2,..., N_k\}$ deployed in the target area. Different types of sensor nodes perceive different types of information such as picture, sound, and odor. Each data type has a weight value according to its importance $\{W_1, W_2,..., W_k\}$. While some of these sensor nodes are identical, their $W$ values are also the same accordingly. Moreover, quality and resolution of the data are considered as factors of credibility. For instance, images taken by cameras with different resolutions and images with different brightness and contrast levels.

Let us define the credibility for the $i^{th}$ event as
\begin{equation}
C_i=\lambda \times W_i,
\end{equation}
where $\lambda$ is an impact factor of the quality and resolution of the sensed data. We consider the maximum $C_i=1$ with $W_i=1$ and $\lambda=1$, meanwhile, we consider the minimum $C_i$ to be equal to 0.


The distance of an animal to the sensor node is considered as a factor of the initial reward. Distance has importance due to two reasons. The first is its effect to the accuracy of the data. For instance, the volume of sounds from the animals may affect the accuracy of classification of the animal species. Another example may be the resolution of the image taken by a sensor node or the distance of the image. The second reason is sensed event with lower distance may help finding the exact location of animal and direct observation by the UAV.

\begin{equation}
I_{dist}=\alpha \times \frac{1}{A_{est}}
\label{Eq:Dist}
\end{equation}
By Equation~\ref{Eq:Dist}, the distance to the animal will be reflected into an estimation of the size of the area where the animal is located. The smaller the estimated area $A_{est}$ is, the more credible the sensing result is. $\alpha$ is a constant value used for adjusting the value $I_{dist}$

We define the duration parameter $I_{dura}$ of the sensed event for the initial reward. Events having longer durations are considered more effective evidences for the animals¡¯ appearances. For instance, an event with a longer duration may infer that the animal prefers staying in the proximity of the sensor node. $I_{dura}$ also used for data types such that when an animal is detected, sensor nodes records series of images with a previously defined frequency. Some environmental noise may also cause similar results. We consider a threshold value $T_{max}$ for the maximum duration that leads to ideal $I_{dura}$ and define the duration parameter as
\begin{equation}
I_{dura}=\frac{T}{T_{max}},
\end{equation}
where $T$ is the duration of the event and $ 0 < I_{dura} \leq 1$.

Finally, we define the initial reward $IR_i$ of the event $i$ as below.
\begin{equation}
\label{eq:IR}
IR_i= \sigma \times C_i\times I_{dist}\times I_{dura},
\end{equation}
where $\sigma$ is a parameter that depends on the type of animal. For instance, information sensed by endangered animal species may have better reward.

With the $IR$, specific event's value of information $VoI_i$ can be calculated. $VoI_i$ shows the value of the sensed event $i$ at the time when it is sent to a mobile sink (UAV). $VoI_i$ has the maximum value at the moment the event is detected and then it gradually decreases as time passes.

\begin{equation}
A_i=I\times R_i
\end{equation}

\begin{equation}
VoI_i=A_i\times e^{-B_it}
\end{equation}
where B is a factor to control the convergence speed of the VoI. Our main objective is to maximize VoI collected by the UAV during its operation.

\section{Animal Monitoring System}
\label{sec:Network Model}

\subsection{Network model}

Considering a large observation area with animals living in it, the goal is to monitor specific animals' appearance. First, a set of sensor nodes with monitoring functionalities are deployed and one UAV is introduced to collect data from deployed sensors. One way about collecting data by the UAV is visiting each sensor directly. However, visiting each sensor node becomes impractical when number of sensors is large. So, we divide the whole area into virtual grids and one cluster head is selected.

\begin{figure}[h]
\centering\includegraphics[width=0.4\textwidth]{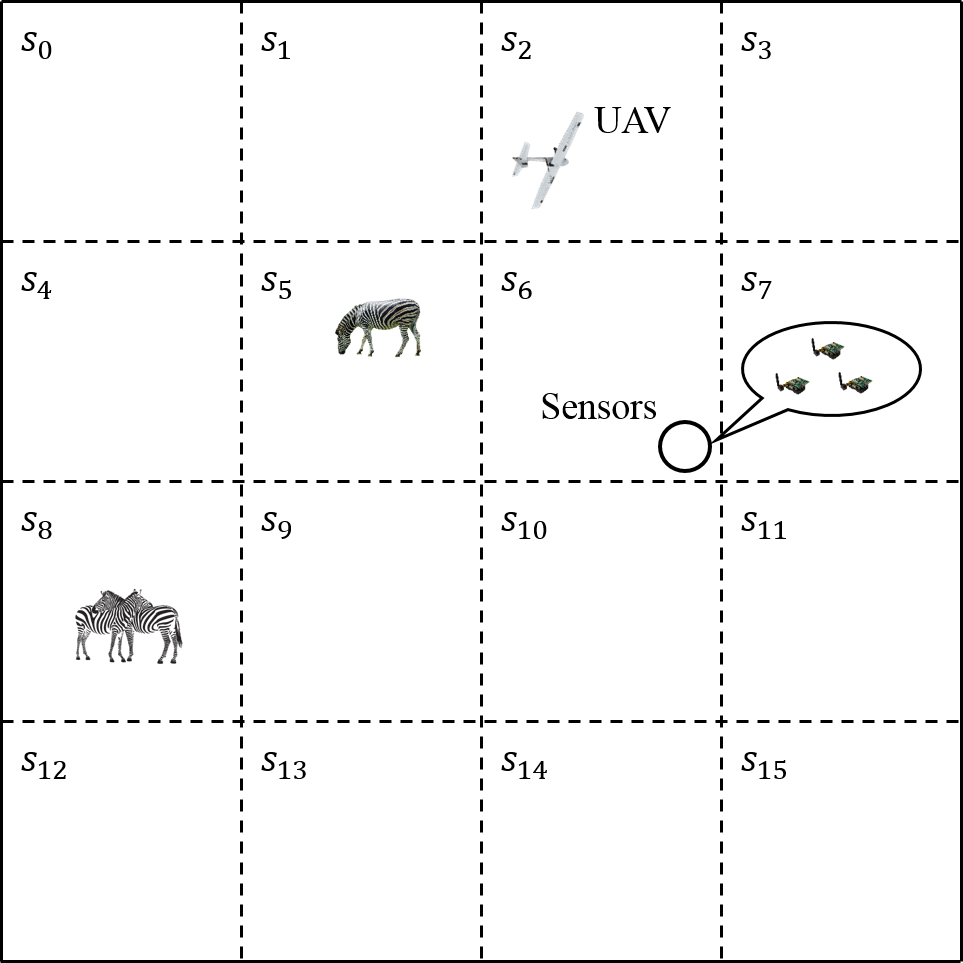}
\caption{The network model.}\label{Fig:NetworkModel}
\end{figure}
 \vspace{4pt}

\subsubsection{Sensor nodes} Sensor nodes are deployed by uniform random distribution in strategic parts of the observation area. As illustrated in Fig.~\ref{Fig:NetworkModel}, sensors in a grid can be treated as a cluster and a cluster head is selected periodically. Sensors inside a virtual grid can communicate with the cluster head directly or via a hop-by-hop communication. Cluster head is responsible for receiving event messages from other sensors and then submitting all the event messages to the UAV when it comes. While routing and clustering are not in the scope of this paper, they are well-investigated problems and there exist various efficient mechanisms~\cite{6672029} ~\cite{5233841}. For instance, Hamidreza et al.~\cite{6672029} propose a cluster head (rendezvous points) selection method by jointly considering node degree and hop distance such that minimizing energy consumption and improving load balance. They also use virtual rendezvous points to increase the performance. In our application, cluster head selection is based on an energy balancing policy which is proposed in ~\cite{5233841}.

\subsubsection{UAV} UAVs have been widely used in various applications due to their several advantages such as flexibility, fast speed and good endurance. In this WSN application, UAV is used as an autonomous mobile sink for gathering time sensitive information. Apart from previous mentioned advantages, using UAV for animal monitoring not only overcomes the geographical challenge but also executes no effects on animals.


Assumptions of the network model are given as follows:\par

\begin{itemize}
\item[-] There is a single UAV as it is the expensive element of the network.
\item[-] The UAV has no energy constraints, while it is not the case for the sensor nodes.
\item[-] The UAV flies with a fixed speed and it only communicates with cluster heads.
\end{itemize}


\subsection{Markov decision-based path planning}

\begin{figure}[h]
\centering\includegraphics[width=0.45\textwidth]{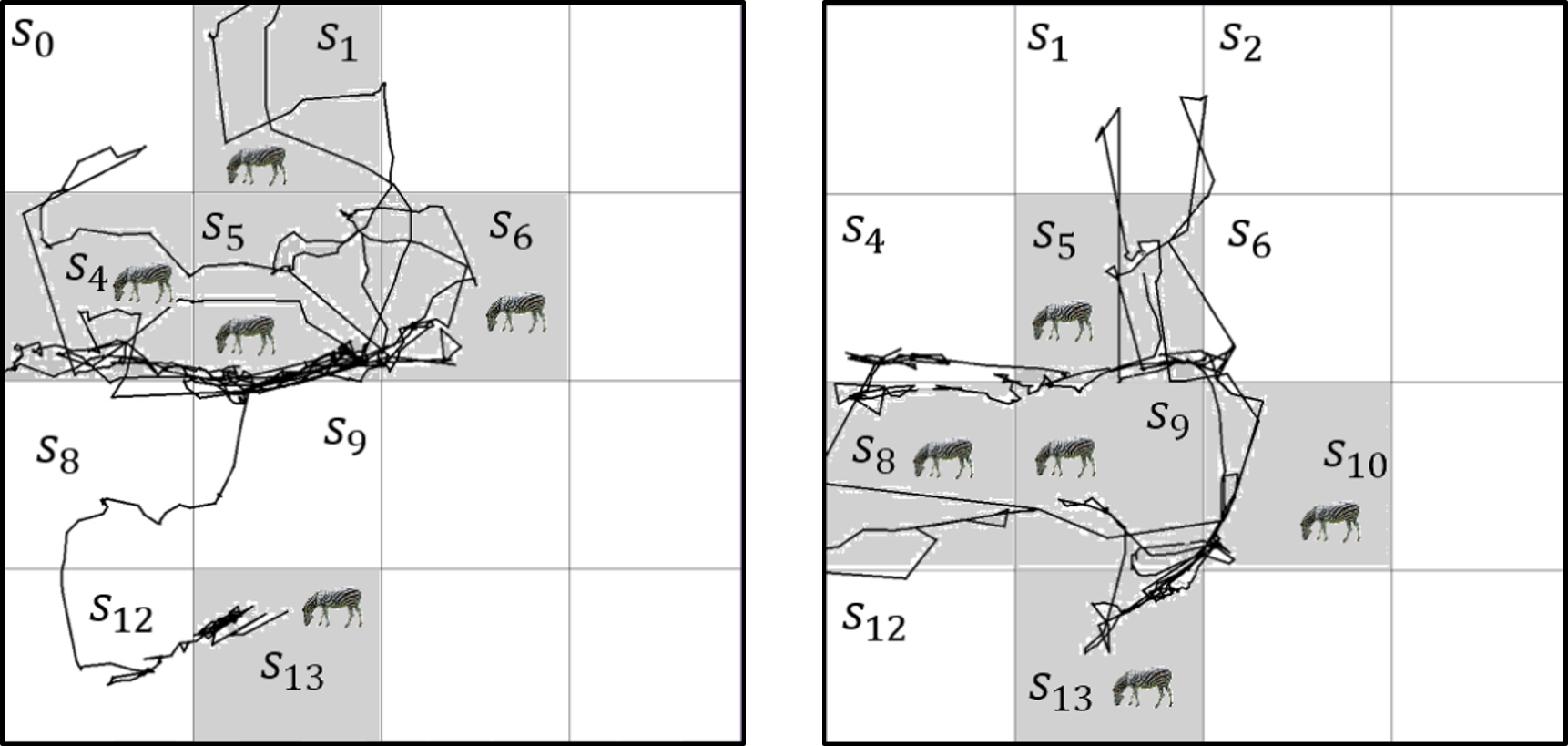}
\caption{Movement trajectories of 2 zebras from the ZebraNet dataset.}\label{Fig:ZebraTraces}
\end{figure}

Wildlife animals have their own habitats, which means they are more likely to stay at a certain location for rest or just having activities in a small area. Fig.~\ref{Fig:ZebraTraces} shows real movement trajectories of 2 zebras in 3 days. As it can be seen in this figure, zebras' mobility choices are not random. Instead, most of the time they prefer to stay in a place or just move nearby. Furthermore, we observe that a zebra can visit an already visited location multiple times. Therefore, sensing a zebra in a region may infer the possibility of future visits in the same region.

Based on this observation, we use a Markov decision process (MDP) model for the path planning problem of UAV. A finite state MDP is a 6-tuple $(S,A,P,D,R,\gamma)$ where $S={s_0, s_1,...,s_m}$ is a finite set of states. $A$ is a set of actions and $P$ is a set of state transition probabilities. $D$ is the initial-state distribution, from which the start state $s_0$ is drawn. $R : S \longrightarrow A$ is the reward function and $\gamma \in [0, 1)$ is a discount factor. In the context of UAV path planning, we define the elements of MDP as follows.

\begin{itemize}
\item[-] \emph{S} is the set of states (grids) in the network.
\item[-] \emph{A} is the set of cardinal directions that UAV can go plus staying in the same grid: \{north, east, south, west, northeast, southeast, southwest, northwest, stay\} .
\item[-] \emph{P} is the set of state transition probabilities. Our model is deterministic, i.e. the probability of ending up in the desired grid (say north neighboring grid) by taking the corresponding action (north) is 1. The probability of accidentally appearing in non-desired states is 0.
\item[-] \emph{D} is the initial-state distribution which is 1 for the top-left corner grid in the network and 0 for all other grids. This means that the algorithm always starts from state $s_0$.
\item[-] \emph{R} is the reward UAV gets if it enters gird $s$. Reward is calculated using Equation~\ref{eq:IR}.
\item[-] $\gamma \in [0,1]$ is the discount factor which incorporates the fact into our model that obtaining the information in distant future worths less compared to getting informed in near future.
\end{itemize}\par \vspace{4pt}

A solution to the MDP can be achieved by repeating the bellman optimality equation~\cite{bellman1957markovian}:

\begin{equation}
V^{\pi}(s):=R_a(s,s')+\gamma\sum_{s'\in S}P_a(s,s')V^{\pi}(s')
\end{equation}
in which  $V^{\pi}(s)$ is the state value function that helps in choosing the best action $a$ according to the policy $\pi$ which leads to state $s'$. If $\pi$ is the optimal policy, then we can reach the optimal value function $V^*(s)$ which gives the best available solution to the MDP:

\begin{equation}
V^*(s):=\max_{a\in A}\{R_a(s,s')+\gamma\sum_{s'\in S}P_a(s,s')V^*(s')\}
\end{equation}
This equation basically describes the highest expected discounted reward for taking action $a$ from state $s$ and following the policy $\pi$ onwards. Therefore, policy $\pi$ can be achieved by iteratively solving the bellman optimality equation and updating the state values.

We use this MDP model to find a solution to the path planning problem in our application. In our model, the set of cluster heads in the network are represented by the set of states $S$ in MDP. In this setting, UAV needs to make decision on its next visiting location after collecting one cluster head's event messages, which can be represented by the state transitions in the MDP model. After visiting each cluster head, the UAV will update its information of the network and decide the next visiting cluster head. This decision process actually have an analogy to the MDP in which rewards of previous actions will have an effect on the next state transition decision. Lastly, maximizing the VoI in our problem actually means optimizing rewards in MDP.

Solving the MDP gives the optimal policy which specifies the best next grid in the observation area that the UAV should visit. To solve the MDP, we use Q-learning algorithm~\cite{watkins1992q}. By utilizing value iteration method, we calculate the Q-value for each state-action pairs using the following equation.

\begin{equation}
\label{Eq:Qvalue}
Q(s,a) \leftarrow R(s,a) + \gamma \max_{a'} Q(s',a')
\end{equation}

\begin{equation}
\label{Eq:reward}
    R(s,a)=
   \begin{cases}
   \sum_{i\in events}IR_i &\mbox{if grid has animals}\\
   R_{Negative} &\mbox{if grid does not have animals}
   \end{cases}
\end{equation}

$Q(s,a)$ is the new Q-value for taking action $a$ when the UAV is in state $s$. $R(s)$ is calculated using Equation~\ref{eq:IR}. $s'$ is the grid the UAV appears in when it takes action $a$ and $a'$ is any action possible when we are in state $s'$. This means each time $a$ is taken from state $s$, UAV reaches state $s'$, calculates the maximum Q-value of next state by taking any arbitrary action from state $s'$, sums it with initial reward it gets as it enters state $s'$, and uses this value to update $Q(s,a)$.

In the context of UAV path planning in this animal monitoring application, when the UAV goes to one of the neighboring grids, it gets some experience (finds some animals, or does not find anything there), it will update the reward of coming to this state from the previous state. This helps the UAV to use these experiences to decide about which grid is best to go from the current grid. Formally, this policy is obtained using the following equation.



\begin{equation}
\pi(s)= \arg\max_a Q(s,a)
\end{equation}

\begin{figure}
\centering\includegraphics[width=0.40\textwidth]{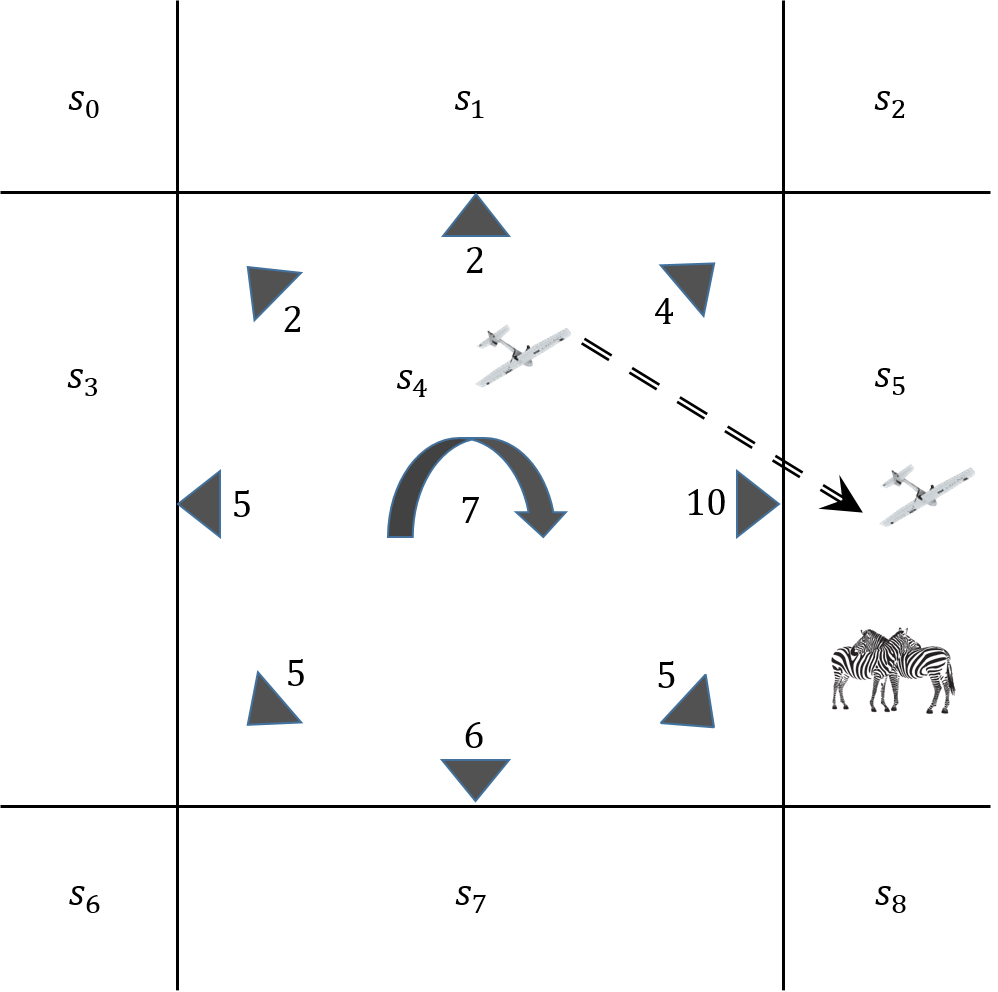}
\caption{Deciding the next state based on Q-values.}\label{Fig:PathPlanning}
\end{figure}

Fig.~\ref{Fig:PathPlanning} illustrates an example of path planning decision for the UAV based on the Q-value of each action in the current state. The UAV is initially in state $s_4$ as it decides which of the adjacent states it should go. The actions are shown by the arrows. Available actions are either visiting the neighboring states or staying in the same state. The Q-value corresponding to each action is shown near each arrow. UAV decides to take action which leads to state $s_5$ since the Q-value of the corresponding action is higher than others. In addition, there is a probability of random exploration during the next state selection process. The exploration is based on the $\epsilon\text{-}greedy$ scenario in Q-learning.


\begin{algorithm}
\caption{Path planning of UAV}
\label{Algpathplanning}
\begin{algorithmic} [1]
\STATE Initialize the UAV in state $s$, with possible $Action [$north, east, south, west, northeast, southeast, southwest, northwest, stay$]$ and their corresponding Q-values $Q(s,a), a\in Action []$. $\epsilon$  is the probability of taking a random action, $0\leq\epsilon\leq1$
\WHILE{Termination condition not reached}
\STATE Generate $RandomNumber \leftarrow Random(0,1)$
\IF{$RandomNumber \leq \epsilon$} 
    \STATE $NextAction = Action[Random(Action.len)]$
\ENDIF
\IF{$RandomNumber > \epsilon$} 
    \IF {all $Q(s,a)$ have the same value}
        \STATE $NextAction = Action[Random(Action.len)]$
    \ELSE
        \STATE $NextAction = Action[\max_{a} Q(s,a).index]$
    \ENDIF
\ENDIF
\STATE Take action $NextAction$, get the reward $R(s,a)$

\STATE $Q(s,a) \leftarrow R(s,a) + \gamma \max_{a'} Q(s',a')$
\STATE // Update Q value
\STATE Current state $s \leftarrow s'$
\ENDWHILE
\end{algorithmic}
\end{algorithm}

The pseudocode for the path planning of UAV is given in Algorithm~\ref{Algpathplanning}. The UAV starts from an arbitrary grid location named as state $s$ with 9 possible actions \{north, east, south, west, northeast, southeast, southwest, northwest, stay\}. Note that the grids near the edge of the observation area have less number of possible actions. The parameter $\epsilon$ controls how much exploration the UAV does instead of following the best path. This enables the UAV to explore the whole area to find potential animal information which the system did not know about before. The UAV will take a completely random action if $\epsilon=1$ while choosing the best possible next grid when $\epsilon=0$. To accomplish this, at each path planning decision, it generate a random number $num$, $0\leq num\leq1$. If $num$ is smaller than $\epsilon$, the UAV randomly selects the next grid from the possible actions (lines [2-6]). If $num$ is greater than $\epsilon$, however, the UAV deterministically decides the best possible next grid according to the Q value of each action. In cases where all possible actions have the same Q value, for instance at the beginning of the network performance, the UAV does random action selection (lines [7-13]). When the next grid is decided, the UAV moves to that grid and receives the reward associated with that state which is determined based on whether the animal is observed or not. Having the immediate rewards, the UAV updates the $Q(s,a)$ using equation~\ref{Eq:Qvalue} (lines [14-17]). After this step, the UAV initializes a new path planning decision which includes repeating the process explained above.


\section{Simulation Study}
\label{sec:Simulation Study}


\subsection{Simulation environment}

The proposed WSN application and the UAV path planning approach are tested with simulation experiments. The whole observation area is deployed by sensors and then classified into grids. One node in each grid is selected periodically as the head node. Head node is responsible for collecting messages from other sensors and communicating with the UAV.

\subsubsection {Dataset and UAV}

We test our model with two real world animal datasets. The first one is the ZebraNet~\cite{Juang02} dataset which was used in our previous work~\cite{7366291}. This dataset contains the location information of 5 zebras in June 2005 at a place near Nanyuki, Kenya. The sampling time of the GPS traces is 10 minutes and the total experiment time is 14 days. The second one is a leopard tortoises dataset which contains the movement traces of 10 leopard tortoises in the Kalahari desert. It includes the traces recorded for three months from January to March in 2013. The sampling time of the GPS traces is 60 minutes. The difference between these two datasets is that animals in the ZebraNet roam around in a more broad area. They keep moving most of the time, for instance examples of their movement traces are shown in Fig.~\ref{Fig:ZebraTraces}. On the other hand, animals in the leopard tortoises dataset usually have activities in a small area. In other words, animal movements pattern in the ZebraNet dataset are more complex and unpredictable while animal activities in the leopard tortoises dataset are more stationary and predictable.

We setup our experimental network by getting the animal movement information from both datasets and then converting them into our grid-based area which is $10km \times 10km$. A single UAV is used for event message collection. The UAV is responsible for collecting animal information from sensors. In general, the higher the UAV speed is, the sooner the information can be collected. However, UAVs with high maximum speed (usually above 200km/h) are very expensive and are only allowed for military purpose. UAVs for commercial purpose have lower speed (usually maximum speed of less than 100km/h) and shorter lifetime between charges. Some existing UAVs have been used for sensor data collection, for instance, the Claw 3 from General Atomics Aeronautical Systems company and the Mantis from AeroVironment company. However, most of the high-speed UAVs are only used for military purpose. In our case, we select a Bayraktar mini UAV which has a good endurance of 1 hour and a moderate maximum flight speed of $60 km/h$. The time unit used in the experiments is 1 minute so the UAV speed is set as $1 km/round$. In the case of battery exhausting, we assume that the UAV can be recharged so that it can continue to work.

\subsubsection {Event and reward}

We define some interested {\em events} in the conducted experiments. The purpose of the application is to detect and predict the appearances of animals. During the operation, when an animal first appears in an area, the sensor nodes can record it. However, if the same animal stays in that small area for a very long time, such as the case when the animal sleeps there, this is valueless to our application and may cause excessive energy consumption to the sensor nodes. Therefore, we define the {\em events} as following.

\begin{itemize}
\item[-] If an animal moves from one grid to another, we count it as an event.\par
\item[-] If an animal always stays in a grid, instead of recording it in every sampling time interval, we periodically record its location once in every $\Delta t$ amount of time.
\end{itemize}


In Section~\ref{sec:Network Model}, equation~\ref{Eq:reward} gives the rewards details when the UAV enters a grid. We discussed the initial reward (IR) in Section~\ref{sec:Mathematical Model}. Each event has an initial reward which is related to the type of animal $\sigma$, the credibility of sensed information $C_i$, the distance $I_{dist}$ and the duration time $I_{dura}$. In our experiment, we assume homogeneous events for simplicity such that we can focus on the UAV's path planning performance. Also, with the homogeneous events assumption, we set the $A_i$ and $B_i$ in VoI to fixed values.
The reason we do this assumption is that these parameters only affect the final specific values of the experimental performance. In other words, specific values of these parameters do not affect the final performance when compared with other approaches. In different problems, these parameters can be set according to the nature of the application.

\subsubsection {Metrics}

To quantitatively evaluate the performance of a path planning strategy, we report the results according to three performance metrics in our simulation study.

\begin{itemize}
\item[-] \emph{Value of information (VoI)}. VoI is the main metric in this animal monitoring application and maximizing it is the primary goal in designing the proposed MDP path planning approach. The definition of VoI is given in Section~\ref{sec:Mathematical Model} and the parameter values used in the experiment is given in Table~\ref{tab:SimulationParameters}.
\item[-] \emph{Time delay}. Since event messages need to be kept in sensors buffer until being sent to the UAV, message delay is an important metric. Long message delays make the event messages useless and as their information content becomes valueless. In this experiment, we measure both the average message delay and the message delay distribution for all of the events.
\item[-] \emph{Number of animals encountered}. As the UAV flies throughout the network, we define a radius $r$ inside which the zebras can be directly observed by the UAV. Although the number of zebras encountered is not the main goal in designing our path planning approach, direct encounters may be more helpful for the monitoring as the UAV can capture higher resolution images.
\end{itemize}\par \vspace{4pt}

\begin{table}[h]
   \centering
\caption{Experimental parameters}

\begin{tabular}{|c|c|}
\hline
Network size & $100km \times 100km$  \\
Time unit (1 round) & 60s  \\
UAV speed & $1 km/round$ \\
VoI parameters ($A_i$, $B_i$) & (10.0, 0.02) \\
Initial reward ($IR_i$)& 10.0 \\
 $\sigma$ & 10.0 \\
 $C_i$ & 1.0\\
 $I_{dist}$ & 1.0\\
 $I_{duration}$ & 1.0\\
Penalty $R_{Negative}$ & -1\\
Radius $r$ for direct observation & $200m$ \\

\hline
\end{tabular}
\label{tab:SimulationParameters}
\end{table}

Table~\ref{tab:SimulationParameters} includes the parameter values used in our experiments. For the performance evaluation of the proposed Markov decision process-based (MDP) path planning approach, we compare its outcomes with three other approaches: greedy, traveling salesman problem-based (TSP) and random. Each of the these approaches independently does path planning for the UAV in order to efficiently collect event messages.

In greedy approach, when the UAV visits a grid, IRs of events in this grid are summed and used for future grid selection. According to greedy approach, the UAV always pursues the highest local IR in its movement. In other words, the UAV always flies to a neighbor grid with the highest IR. In the case of multiple neighbor grids having the same IRs, the UAV randomly selects one of them. In TSP approach, solution to the TSP problem is used as it provides the shortest path of visiting all grids among all these possible paths. Given the network structure, the path produced by TSP is actually fixed, which means the UAV always flies along a pre-set movement path to ensure visiting all the nodes. The fixed path feature of TSP results in a very stable experimental performance and can also be used as a good reference for comparing with adaptive approaches. Lastly, in random approach, UAV simply selects its next destination grid randomly among all the grids in the network.


\subsection{Simulation results}
This section reports the experimental results of the compared path planning approaches. We implement 4 independent UAVs which are controlled by MDP, greedy, TSP and random approaches respectively. The results are obtained from an average of 10 simulation runs. To clearly distinguish the experimental results from two different datasets, each figure is marked as first dataset or second dataset in the caption. Since the total experimental time in both datasets is too long, we show the experimental result of the first 70 hours in the first dataset which contains 5682 GPS traces, while in the second dataset we show the experimental result of the first 200 hours which contains 7896 GPS traces.

\begin{figure}[htbp]
\centering\includegraphics[width=0.45\textwidth]{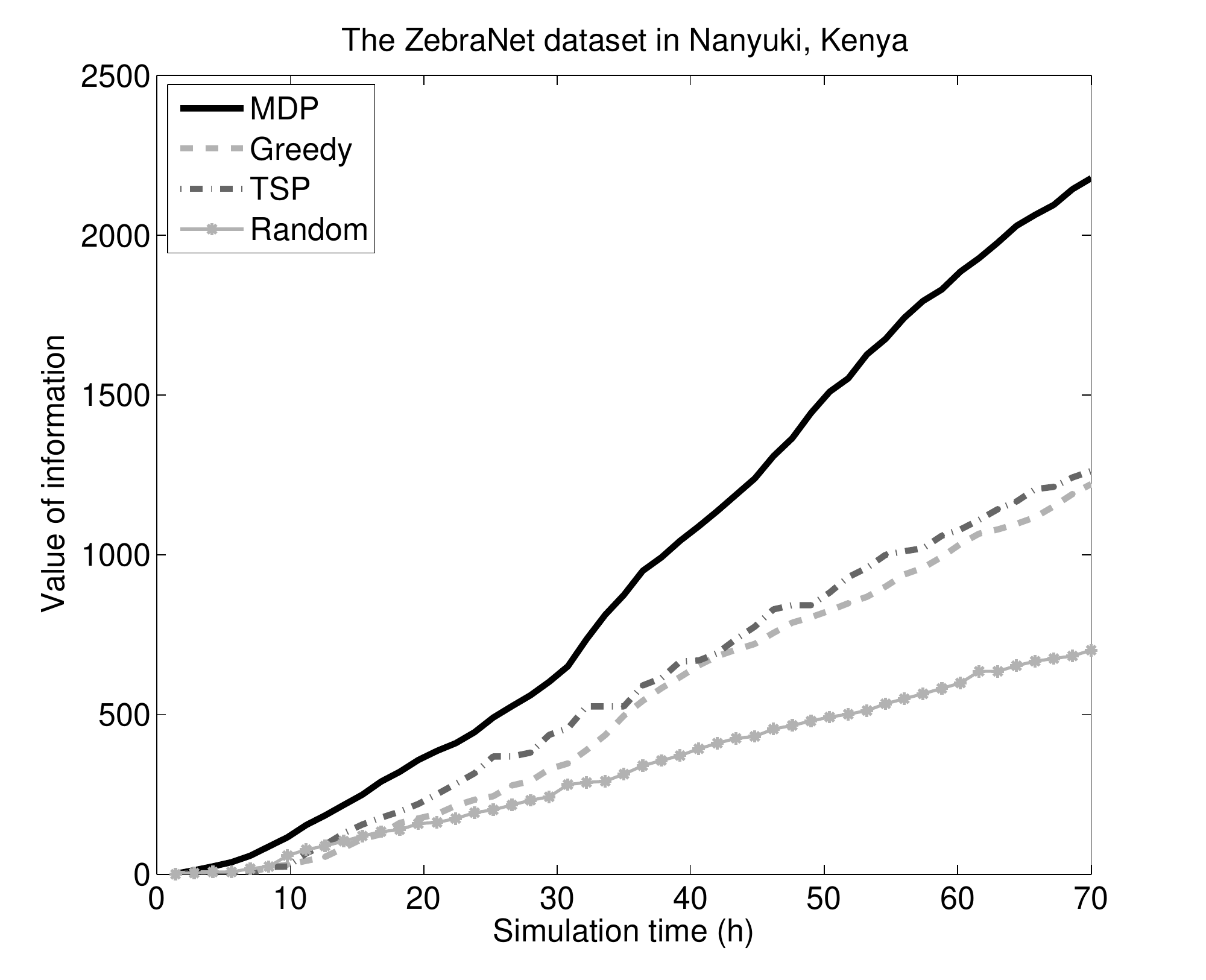}
\caption{ZebraNet dataset, performance comparison of VoI by simulation time, with network settings: $4\times4$ grids, $\epsilon=0.2$}\label{Fig:VoIComp}
\end{figure}

\begin{figure}[htbp]
\centering\includegraphics[width=0.45\textwidth]{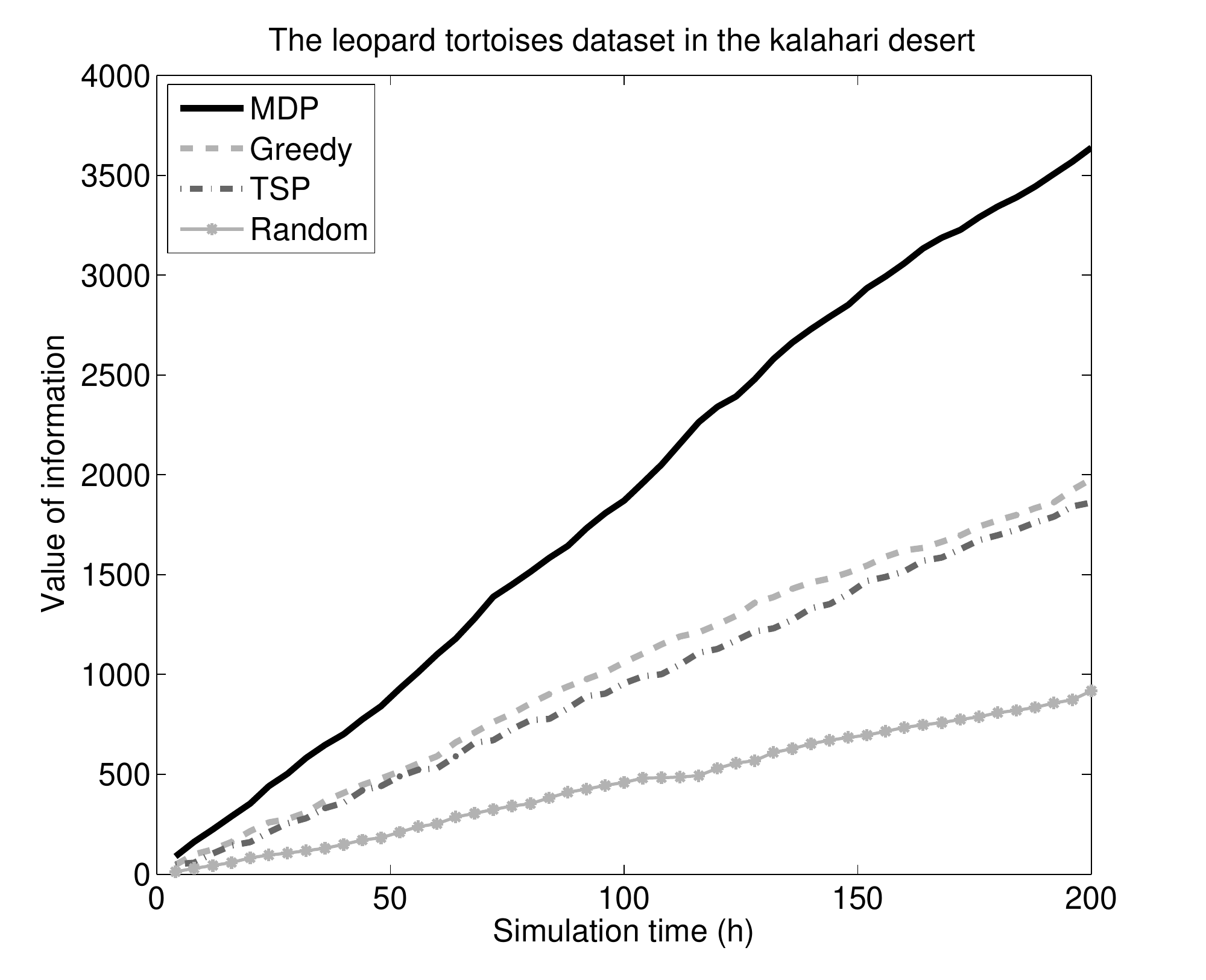}
\caption{Leopard tortoises dataset, performance comparison of VoI by simulation time, with network settings: $4\times4$ grids, $\epsilon=0.2$}\label{Fig:NewVoIcompare}
\end{figure}

Fig.~\ref{Fig:VoIComp} and Fig.~\ref{Fig:NewVoIcompare} show the performances of the path planning approaches in terms of the VoI by simulation time. The network structure is set to $4\times4$ grids. In MDP, the $\epsilon=0.2$ means the UAV has a $20\%$ chance of randomly exploring the area during path planning while a $80\%$ chance of exploitation of the best available policy. At the beginning of the experiment, the MDP-UAV does not contain any animal appearance information, so it randomly visits each grid. Once the MDP-UAV gets animal appearance information from one grid, it gets a reward from that grid and updates the Q-value table such that it has a higher probability to visit back that grid and its neighbor grids. The $\epsilon\text{-}greedy$ policy is applied at each path planning step such that animal information in other grids can be detected by the UAV. In such a continuous learning way, the MDP-UAV is more likely to go to those areas that previously had animals.

As the learning process continues, the VoI obtained by MDP-UAV gradually outperforms other path planning approaches on both datasets. The random approach performs the worst due to its totally random grid selection. The result of TSP is better than greedy under both datasets. This is because to a given network structure, the TSP-UAV always flies along a fixed path in the network, which means it has a very stable performance. But in greedy approach, the Greedy-UAV keeps go to neighbor grids where it finds animals last time. In other words, the greedy approach is trapped into a local minima and cannot adapt itself based on animal movements. Note that the results of all algorithms are pretty consistent across different datasets.

\begin{figure}[htbp]
\centering\includegraphics[width=0.45\textwidth]{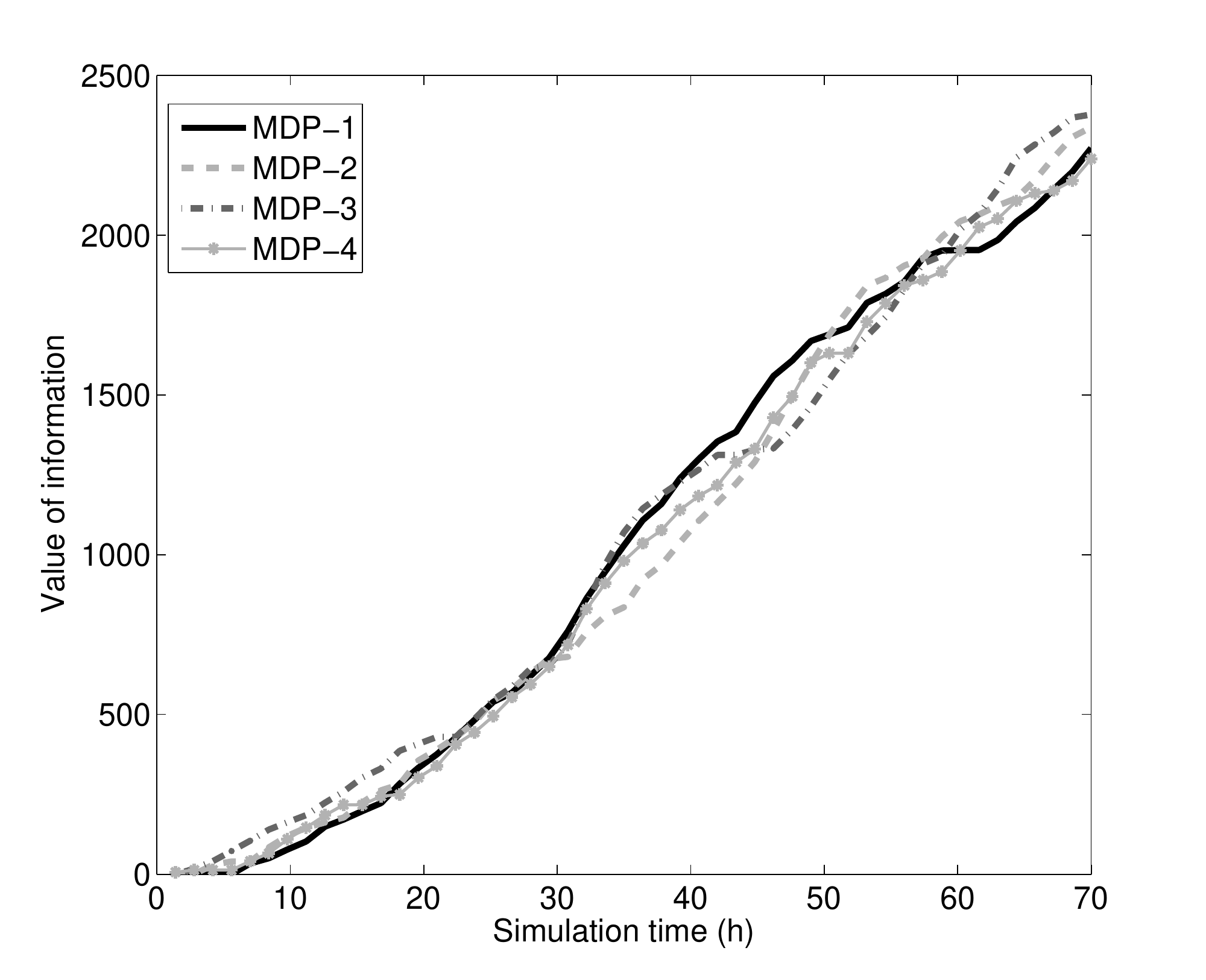}
\caption{ZebraNet dataset, stable performance of MDP, with network settings: $4\times4$ grids, $\epsilon=0.2$}\label{Fig:OldStable}
\end{figure}

\begin{figure}[htbp]
\centering\includegraphics[width=0.45\textwidth]{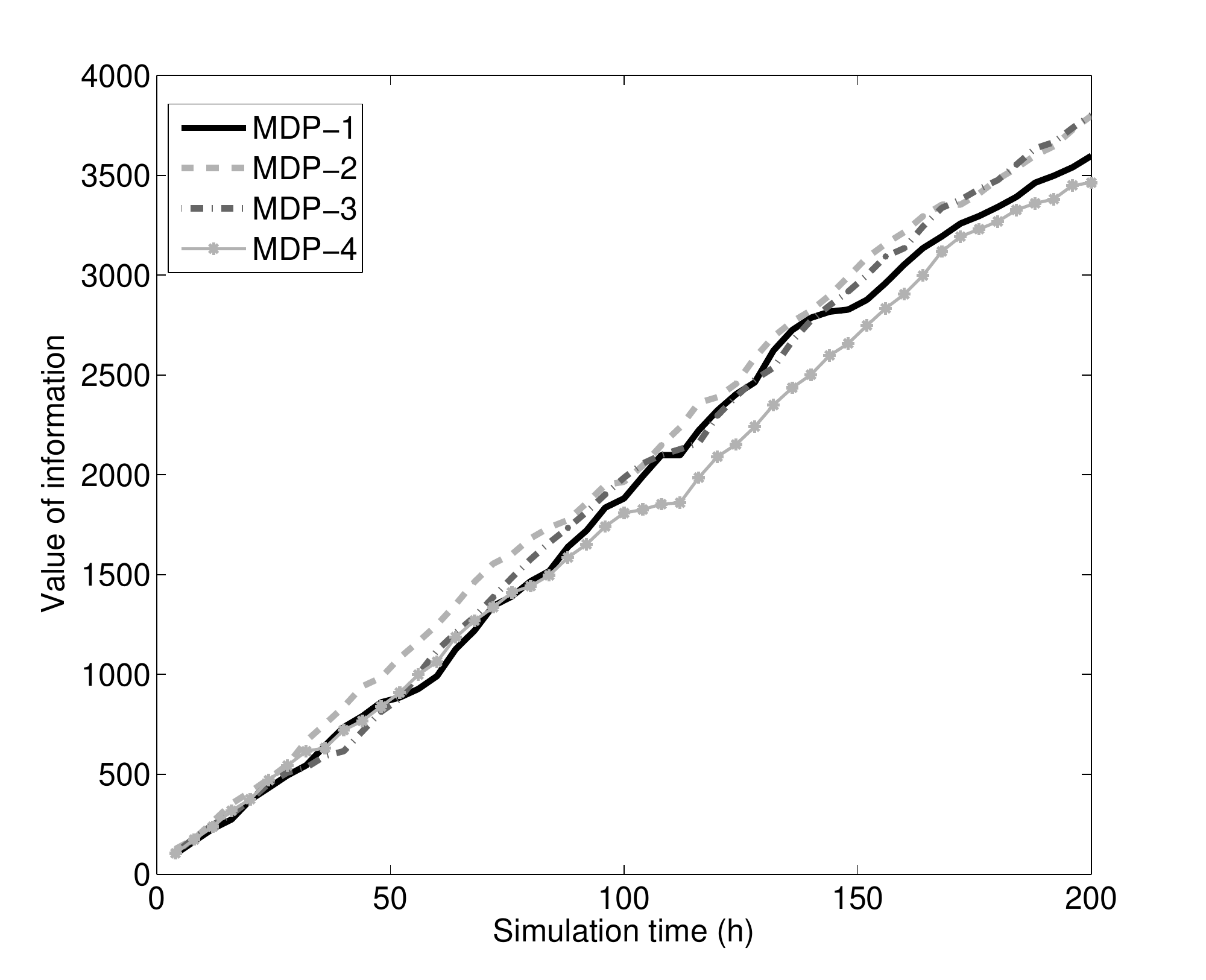}
\caption{Leopard tortoises dataset, stable performance of MDP, with network settings: $4\times4$ grids, $\epsilon=0.2$}\label{Fig:NewStable}
\end{figure}

The performance results of different simulation runs using the MDP approach are shown in Fig.~\ref{Fig:OldStable} and Fig.~\ref{Fig:NewStable}. Note that the $\epsilon\text{-}greedy$ policy brings stochasticity in selecting the next state and therefore leads to randomness in the results. To examine the stability of our approach despite the randomness, four independent experiments of MDP approach with the same parameter settings are conducted on both datasets. It can be seen that though some small differences exist among different experimental runs, the overall outcome is consistent.

\begin{figure}[htbp]
\centering\includegraphics[width=0.45\textwidth]{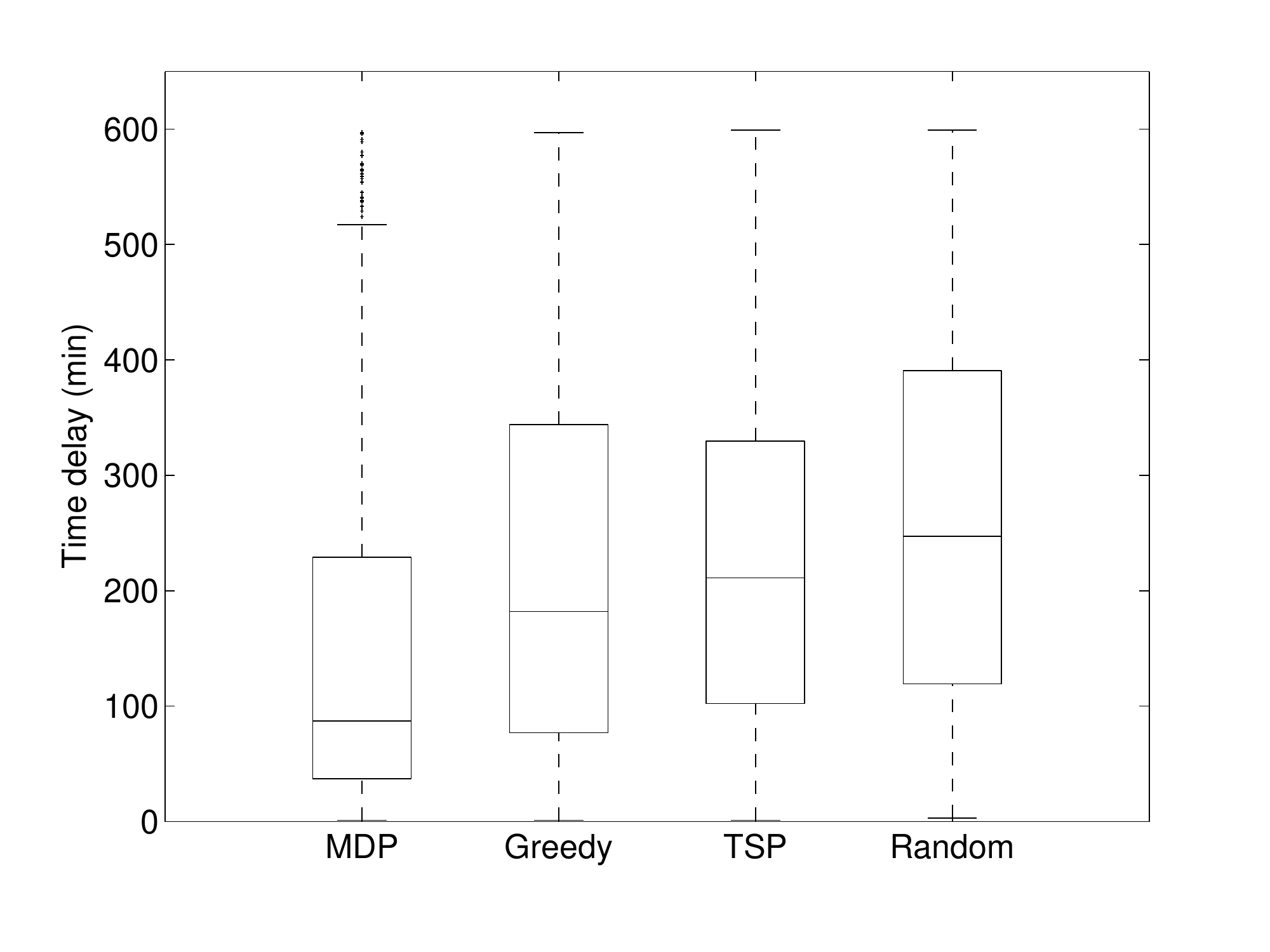}
\caption{ZebraNet dataset, message delay distribution of the MDP, greedy, TSP, and random approaches, with network settings: $4\times4$ grids, $\epsilon=0.2$}\label{Fig:OldtimeDelay}
\end{figure}

\begin{figure}[htbp]
\centering\includegraphics[width=0.45\textwidth]{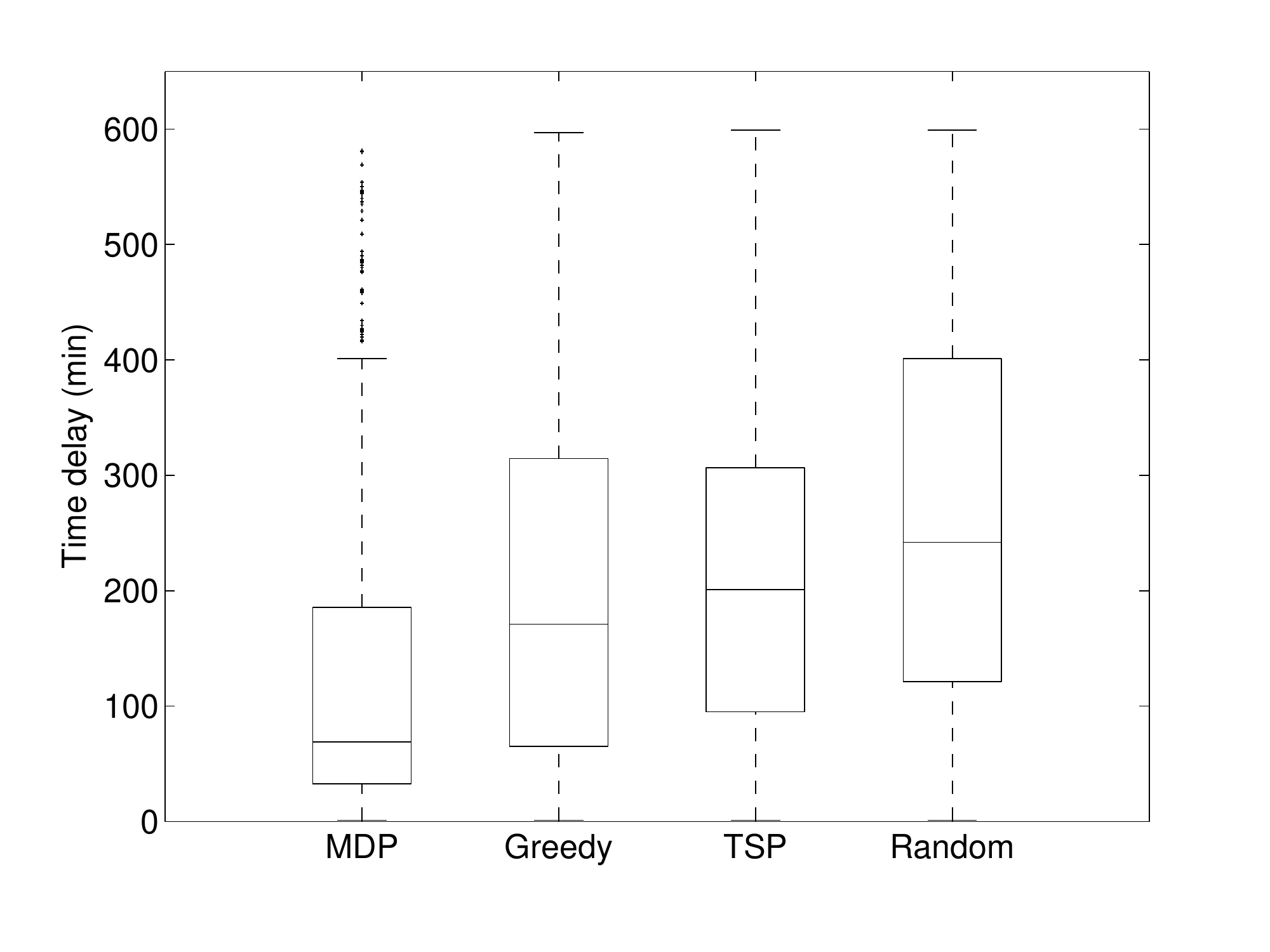}
\caption{Leopard tortoises dataset, message delay distribution of the MDP, greedy, TSP, and random approaches, with network settings: $4\times4$ grids, $\epsilon=0.2$}\label{Fig:NewtimeDelay}
\end{figure}

We include the results for the message delay distribution of both datasets in Fig.~\ref{Fig:OldtimeDelay} and Fig.~\ref{Fig:NewtimeDelay}. It can be seen that the MDP approach has a much more stable message delay distribution compared with other path planning approaches. Due to the continuous learning process, the MDP-UAV keeps visiting those 'hot' areas in the network which enables the sensed messages to be collected in a timely manner. So it has the smallest median and the smallest third quartile among the four approaches. It means 75\% animal events can be collected by the MDP-UAV in a short prediod of time. When compared with the TSP approach which has a median of 200 and the third quartile of 320, the MDP approach turns to be much more effective and intelligent in path planning and animal detecting. These distribution results also support the findings about the value of information shown in Fig.~\ref{Fig:VoIComp} and Fig.~\ref{Fig:NewVoIcompare}. VoI is highly correlated to the time delay, i.e. messages delivered quickly will result in a higher obtained VoI.

\begin{figure}[htbp]
\centering\includegraphics[width=0.45\textwidth]{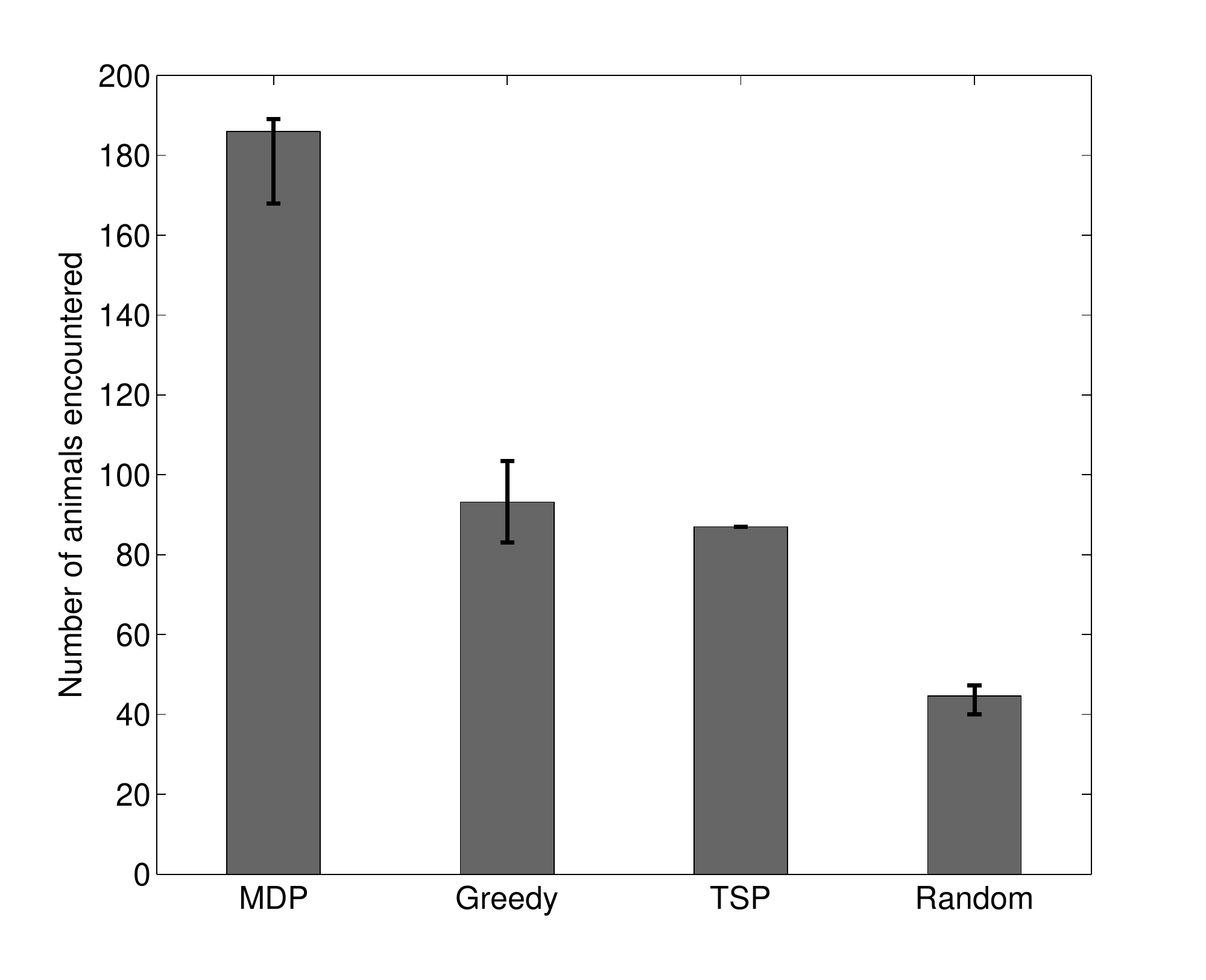}
\caption{ZebraNet dataset, Times of zebras encountered for the MDP, greedy, TSP, and random approaches, with network settings: $4\times4$ grids, $\epsilon=0.2$}\label{Fig:OldZebraEncounter}
\end{figure}

\begin{figure}[htbp]
\centering\includegraphics[width=0.45\textwidth]{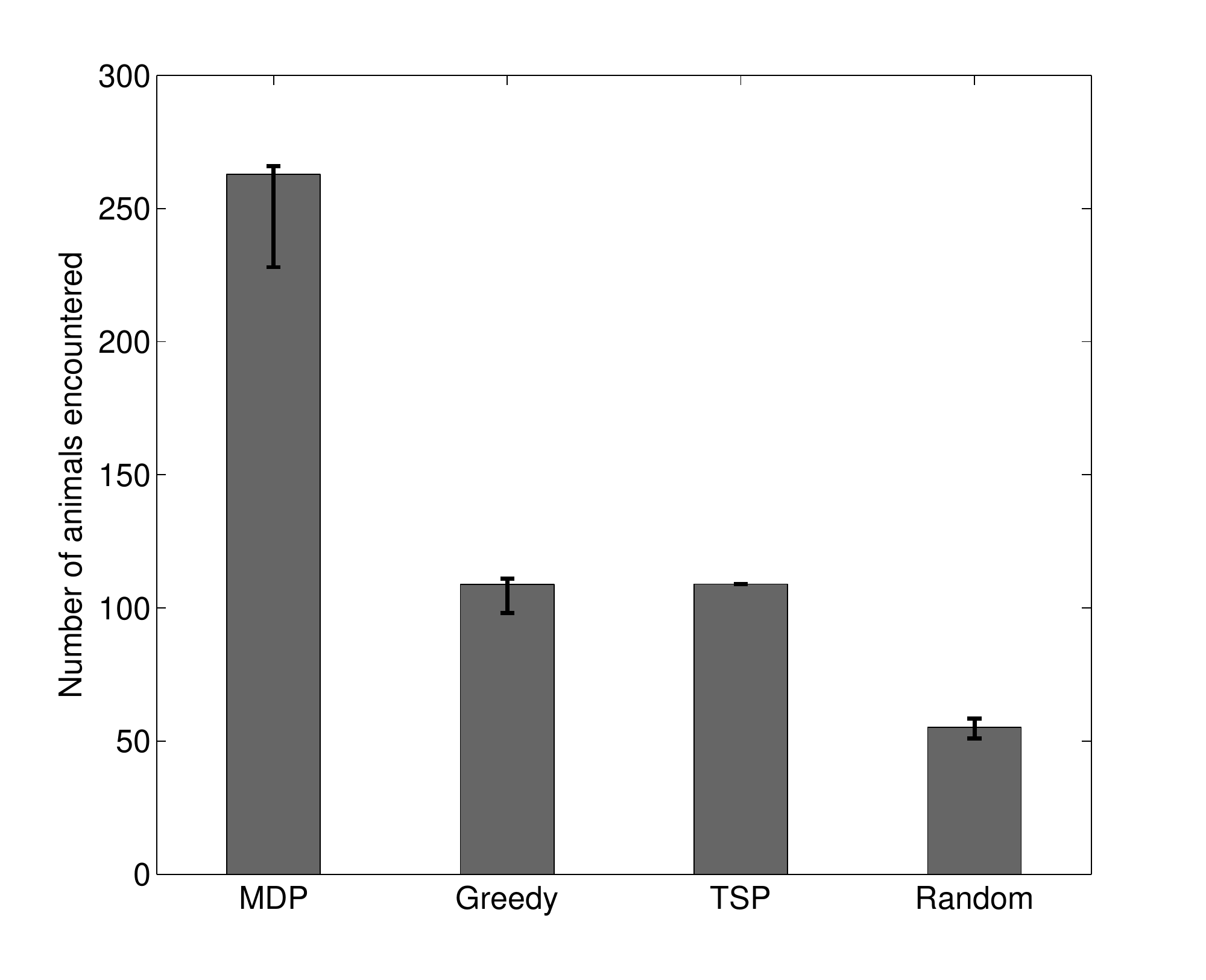}
\caption{Leopard tortoises dataset, Times of zebras encountered for the MDP, greedy, TSP, and random approaches, with network settings: $4\times4$ grids, $\epsilon=0.2$}\label{Fig:NewZebraEncounter}
\end{figure}

We depict the results for the total number of times the UAV encountered animals in Fig.~\ref{Fig:OldZebraEncounter} and Fig.~\ref{Fig:NewZebraEncounter}. In each dataset, the MDP-UAV has encountered animals more often compared to the other approaches. Note that the standard deviation is smallest for the MDP-UAV approach. This finding means that we reached our primary goal of designing this approach which is effective and reliable predicting animal movements. While the greedy and TSP approaches seem to result in the same frequency of encountering animals, TSP turns out to be more stable than greedy. This is because the greedy-UAV visits the grids randomly when the IRs of all neighbor grids are equal to each other.

\begin{figure}[htbp]
\centering\includegraphics[width=0.45\textwidth]{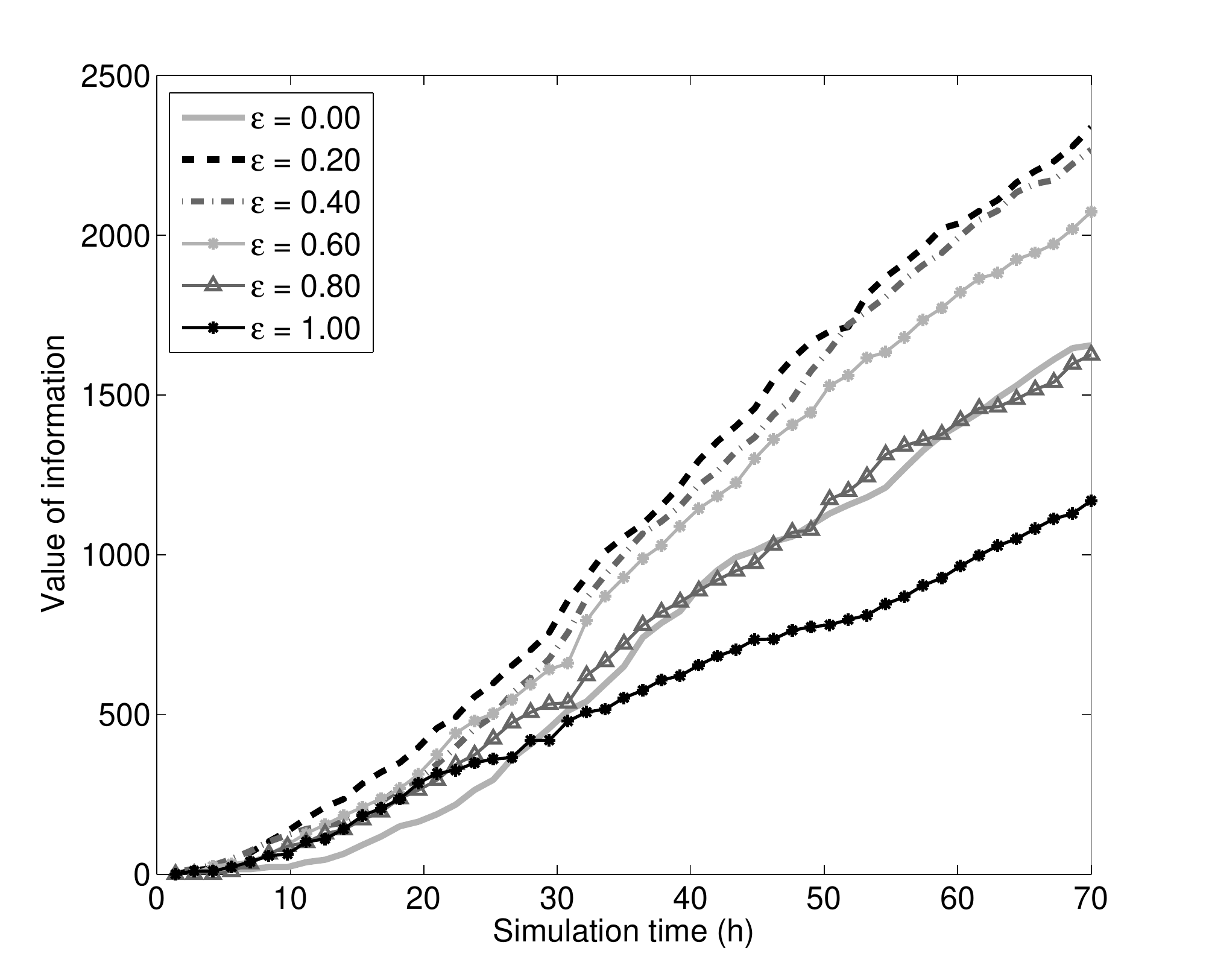}
\caption{ZebraNet dataset, The impact of $\epsilon$ on VoI by simulation time, with network settings: $4\times4$ grids}\label{Fig:OldEpsilon}
\end{figure}

\begin{figure}[htbp]
\centering\includegraphics[width=0.45\textwidth]{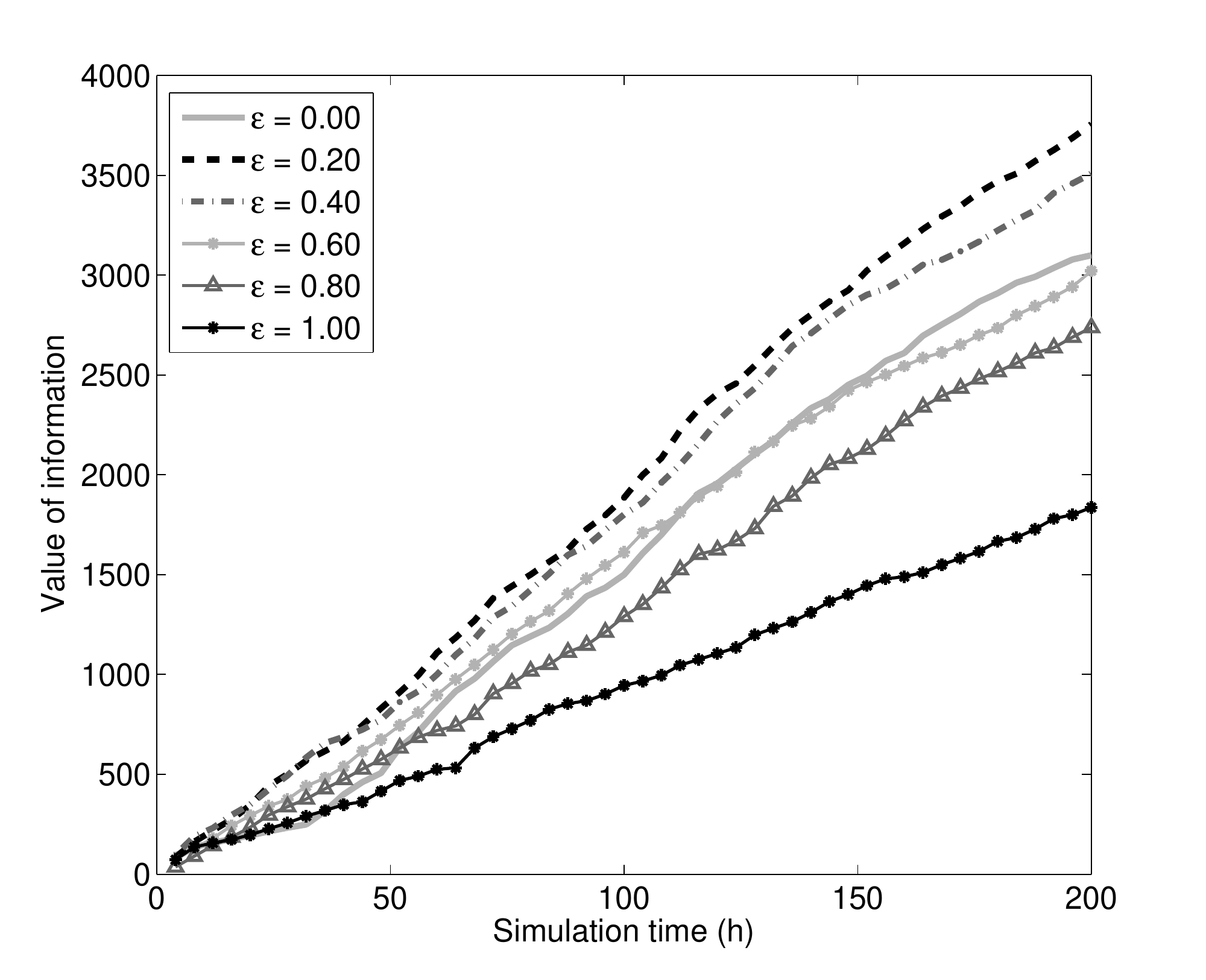}
\caption{Leopard tortoises dataset, The impact of $\epsilon$ on VoI by simulation time, with network settings: $4\times4$ grids}\label{Fig:NewEpsilon}
\end{figure}

Note that all the previous experimental results come with the network settings $4\times4$ grids and $\epsilon=0.2$. In order to investigate the impact of these network parameters on the performance, Fig.~\ref{Fig:OldEpsilon} and Fig.~\ref{Fig:NewEpsilon} show the impact of varying the parameter $\epsilon$ in $\epsilon\text{-}greedy$ approach on the MDP-UAV's VoI performance. $\epsilon$ is the parameter that indicates the probability of stochastic state decisions in MDP, i.e., the probability of random grid selection at each step of path planning. $\epsilon = 0.0$ basically means turing off the exploration feature, i.e. the UAV always does deterministic path planning at each step. On the other hand, $\epsilon = 1.0$ means the UAV does totally stochastic path planning. In this case, the Q-value table does not have any impact on its decisions. The MDP-UAV gets the worst performance when $\epsilon = 1.0$ on both datasets. This is easy to understand because the UAV always performs random selection at each step. As the value of $\epsilon$ decrease, the MDP-UAV's performance gradually increases. The MDP-UAV gets the best performance when $\epsilon = 0.2$ on both datasets. The performance draws back when the $\epsilon$ value continues to decrease. This is because with low $\epsilon$ values, the UAV looses its exploration ability which is important to detect potential animal appearance. It can be understood that when $\epsilon = 0.0$, the MDP-UAV's performance decreases to the level with $\epsilon = 0.8$ and $\epsilon = 0.6$ in the first dataset and second dataset respectively. Empirically, we use $\epsilon = 0.2$ in our experiments as it provides the best trade-off.

\begin{figure}[htbp]
\centering\includegraphics[width=0.45\textwidth]{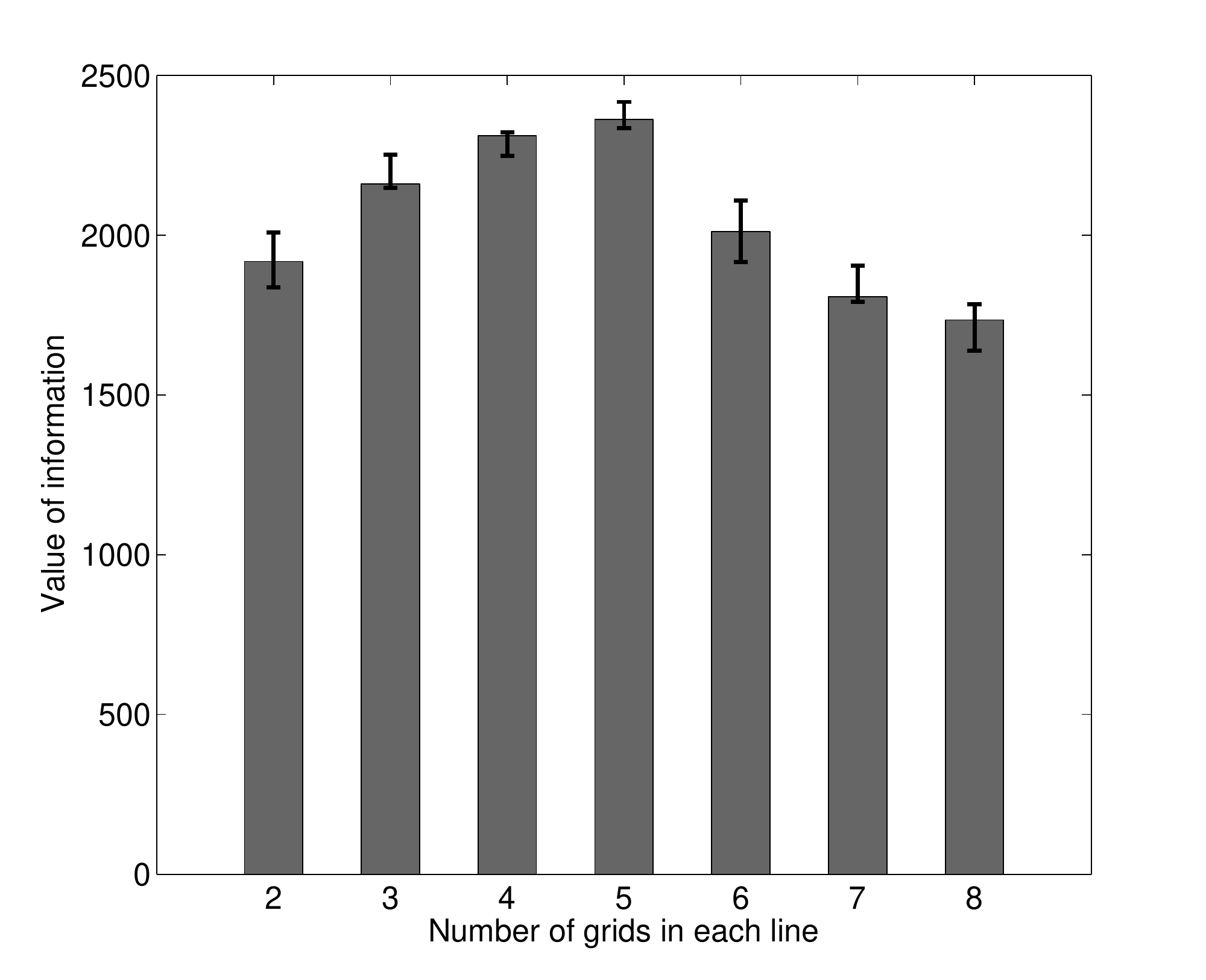}
\caption{ZebraNet dataset, Number of grids in each line, with network settings: $\epsilon=0.2$}\label{Fig:Olddifferentsize}
\end{figure}

\begin{figure}[htbp]
\centering\includegraphics[width=0.45\textwidth]{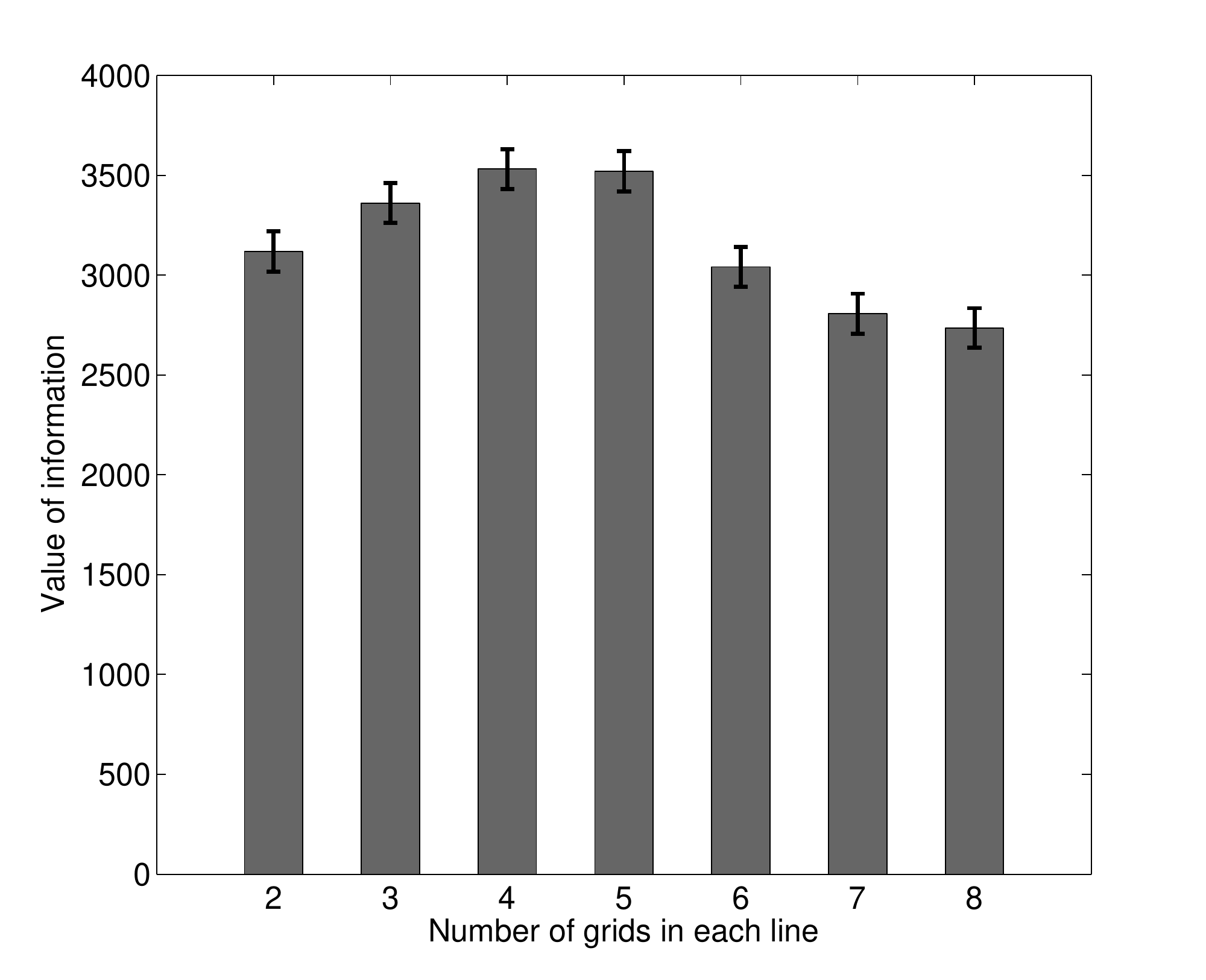}
\caption{Leopard tortoises dataset, Number of grids in each line, with network settings: $\epsilon=0.2$}\label{Fig:Newdifferentsize}
\end{figure}

Fig.~\ref{Fig:Olddifferentsize} and Fig.~\ref{Fig:Newdifferentsize} show the impact of the number of grids on the MDP-UAV's VoI performance. The number of grids is related to the grid size as well as the total number of the MDP states. When the total number of grids is smaller, say less than $3\times3$, it means the area size of each grid is larger. Hence, the UAV needs more time to visit each grid, and it would waste time if there is no event taking place in that grid. On the contrary, as the total number of grids increases, the predication becomes harder. It can be seen from Fig.~\ref{Fig:Olddifferentsize} and Fig.~\ref{Fig:Newdifferentsize} that the best VoI performance can be acquired when the grid number is between $4\times4$ and $5\times5$. Empirically, we use $4\times4$ grids in our experiments as it provides the best performance.

Overall, it can be seen in both experimental results that the MDP approach highly outperforms the other path planning approaches. Compared to the greedy approach as an example, MDP produces a 80\% increase in VoI, much lower median value in message delay, and a 90\% increase in the number of zebras encountered. Although greedy path planning approach also relies on the previous information, it only considers the last time of zebra appearances in each grid. Unlike the greedy approach, the UAV in MDP approach continuously learns about the events occuring throughout the network and then updates the Q-value of each grid. The Q-table maintained by the UAV actually reflects all previous information of zebras activities. As the learning process continues, the UAV is more likely to go to the hot-spots where zebras often appear. Additionally, both exploration and exploitation modes are introduced so that while the UAV is visiting the hot-spots, it also explores new regions in the observation area.


\section{Related Work}
\label{sec:RelatedWorks}

\subsection{Tracking animals with sensor networks}

Tracking animals can be viewed as specific applications of object tracking problems. Its main goal is tracking certain animals in a monitored area and reporting their location and other information to the application��s users~\cite{5676238}. Many animal tracking technologies have been proposed and implemented by engineers and wildlife researchers~\cite{Juang02,6115197,1354585,6093203}. One main technology is the wearable GPS-based animal tracking devices. Juang et al.~\cite{Juang02} present their ZebraNet project in which a low-power wireless system is built for position tracking of zebras. Tracking nodes are installed on zebras and record zebras' GPS positions periodically. In their research, they investigate system design ideas, communication protocols between tracking nodes, and how sensor specifications such as battery lifetime and weight limit the system performance. In addition, some recent research~\cite{7348035,6292789,6533937} on animal behaviour gather animal movement data by installing wearable GPS devices. Although remote sensing can be used for sensing different types of data from a large area~\cite{santos2014investigation,ebrahimi2013use}, sensor networks seem to be a more feasible and reliable choice for animal monitoring.

In recent years, camera sensor networks emerge due to the advancements in hardware technology which provides sufficient bandwidth for transferring video. Camera sensor networks greatly promote wild-life research by providing much more animal related information such as image, sound and video.  He et al.~\cite{7404334} develop integrated camera-sensor networking systems and deploy them at large scales for collaborative wildlife monitoring and tracking. They develop an eMammal cyber infrastructure in which advanced computational and informatics tools are used to analyze and manage massive wildlife monitoring data. Animal species recognition is the problem they tried to solve by using machine learning methods to train a model based on large number of images. Similar studies based on camera sensor networks are conducted in~\cite{6885136} ~\cite{5724879}.



In addition, UAVs are being increasingly used as sensor nodes for monitoring various species in nature. Tuna et al.~\cite{Tuna12} propose to use UAVs for deployment of sensor nodes for post-disaster monitoring. Hodgson et al.~\cite{hodgson2013unmanned} use ScanEagle UAV to survey marine mammals. Their results indicate that UAVs are not limited by sea conditions as sightings from manned surveys. Chamoso et al.~\cite{chamoso2014uavs} propose to use UAVs for scanning large areas of livestock systems. Using visual recognition techniques, the recorded images are used to count and monitor animal species. Akbas et al.~\cite{akbas2012actor} propose the use of aerial sensor networks consisting of UAVs for volcanic eruption monitoring. They propose positioning approaches for multiple UAVs based on the Valence Shell Electron (VSEPR) model of molecular geometries. Our study differs from the aforementioned ones as we propose using UAV as a major element of the WSNs for monitoring purpose.


\subsection{Path planning for mobile sinks in WSNs}

Mobile sinks provide great advantages such as distributing the energy consumption throughout the network and increasing network lifetime. They can be programmed to move to different areas of WSN to accomplish application-specific tasks. The common goal of path planning for mobile sinks is to maximize the information collected while minimizing travel time. In such applications, mobile sinks can either visit each sensor node directly or just visit a subset of them. Hollinger et al.~\cite{6214701} address path planning for autonomous underwater vehicle (AUV) to collect data from an underwater sensor network. Each AUV needs to collect as much data as possible while considering fuel expenditure, i.e., the travel time. Salarian et al.~\cite{6672029} propose a path planning approach for the mobile sink called weighted rendezvous planning (WRP) to collect data from the set of rendezvous points (RPs). WRP assigns weights to sensor nodes based on the number of data packets and hop distance, and selects the sensor nodes with the highest weight as RPs. The goal of WRP is to collect all data within a given deadline while minimizing energy consumption. Cheng et al.~\cite{7332969} propose a Traveling Salesman Problem (TSP) based path planning algorithm for the mobile sink. Instead of visiting each sensor in the observation area, the mobile sink only visits a set of virtual points which are actually overlapping areas of communication ranges of sensors. Their approach first finds out those overlapping areas based on an IRO rule and then assigns contribution value to each overlapping area based on the number of sensors that a mobile sink can communicate in that area. The mobile sink selects visiting points with contribution values and visits them with a TSP strategy such that the whole data can be collected in the shortest traveling path.

Research on mobile sinks to accomplish other tasks is also well conducted. Ma et al.~\cite{4302733} use one SenCar as a mobile sink for data collection in a static sensor network. Their research reveals the effects of traveling path on network lifetime in a given network. They propose a heuristic algorithm for planning the traveling path such that traffic load can be balanced. Basagni et al.~\cite{basagni2014maximizing} investigate the problem of maximizing value of information in underwater sensor networks. They formulate the problem using an Integer Linear Programming (ILP) model for path planning of underwater vehicles. Their method achieves better results in terms of value of information compared to a greedy heuristic. Rahmatizadeh et al.~\cite{rahmatizadeh2014routing} study sink mobility in virtual coordinates domain and propose a routing strategy to minimize energy consumption while notifying the nodes about the latest location of the sink. Solmaz and Turgut~\cite{Solmaz-2014-WINET} propose positioning approaches for multiple mobile sinks to optimize event coverage using WSNs.





\section{Conclusion}
\label{sec:Conclusion}

In this paper, we propose using UAV-aided WSNs for animal monitoring in wildlife areas. Motivated by the movement features of animal, we propose a Markov decision based path planning approach for the UAV. The decision policy is dynamically based on the learning knowledge of animal appearance. The UAV predicts animal¡¯s active areas and visits these predicted areas such that maximizing the overall value of information obtained from the entire network. The proposed network model is evaluated using real-world mobility traces of zebras and leopard tortoises. Simulation results show that the performance of the proposed path planning approach is better than random, greedy, and TSP-based approaches in terms of VoI, message delay and number of directly observed animals. In addition to animals tracking, the proposed model in this research is still effective to other scenarios like tracking objects who have some regular visiting areas, monitoring some dynamic changing sub-areas in a large network. The UAV in our model can find out these dynamic hot areas by exploring the monitored area and learning the obtained knowledge.




\bibliographystyle{IEEEtran}

\bibliography{journalreference}

\parpic{\includegraphics[width=1in,clip,keepaspectratio]{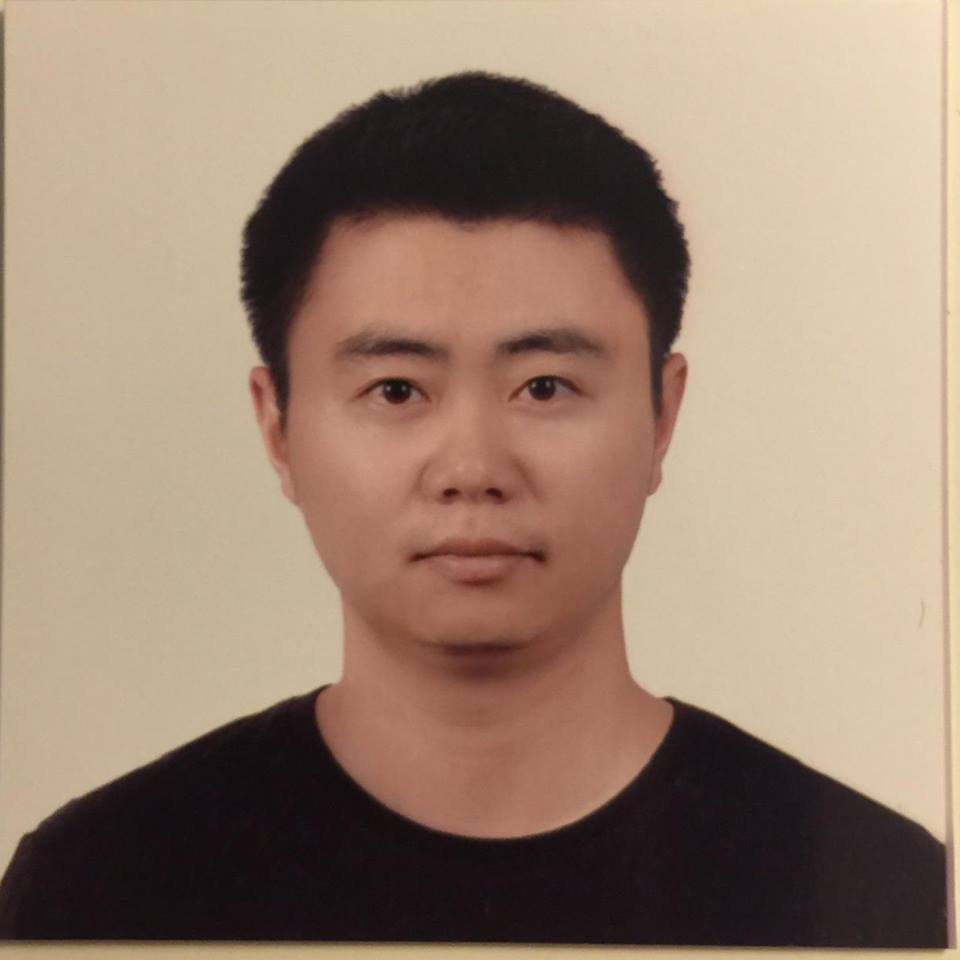}}
\noindent {\bf Jun Xu} is currently working toward the PhD degree in Computer Science from the Department of Electrical Engineering and Computer Science, University of Central Florida (UCF). He received his MS degree in Electrical Engineering from Beijing University of Posts and Telecommunications, China. His research interests include mobile wireless sensor networks and agent path planning.

\parpic{\includegraphics[width=1in,clip,keepaspectratio]{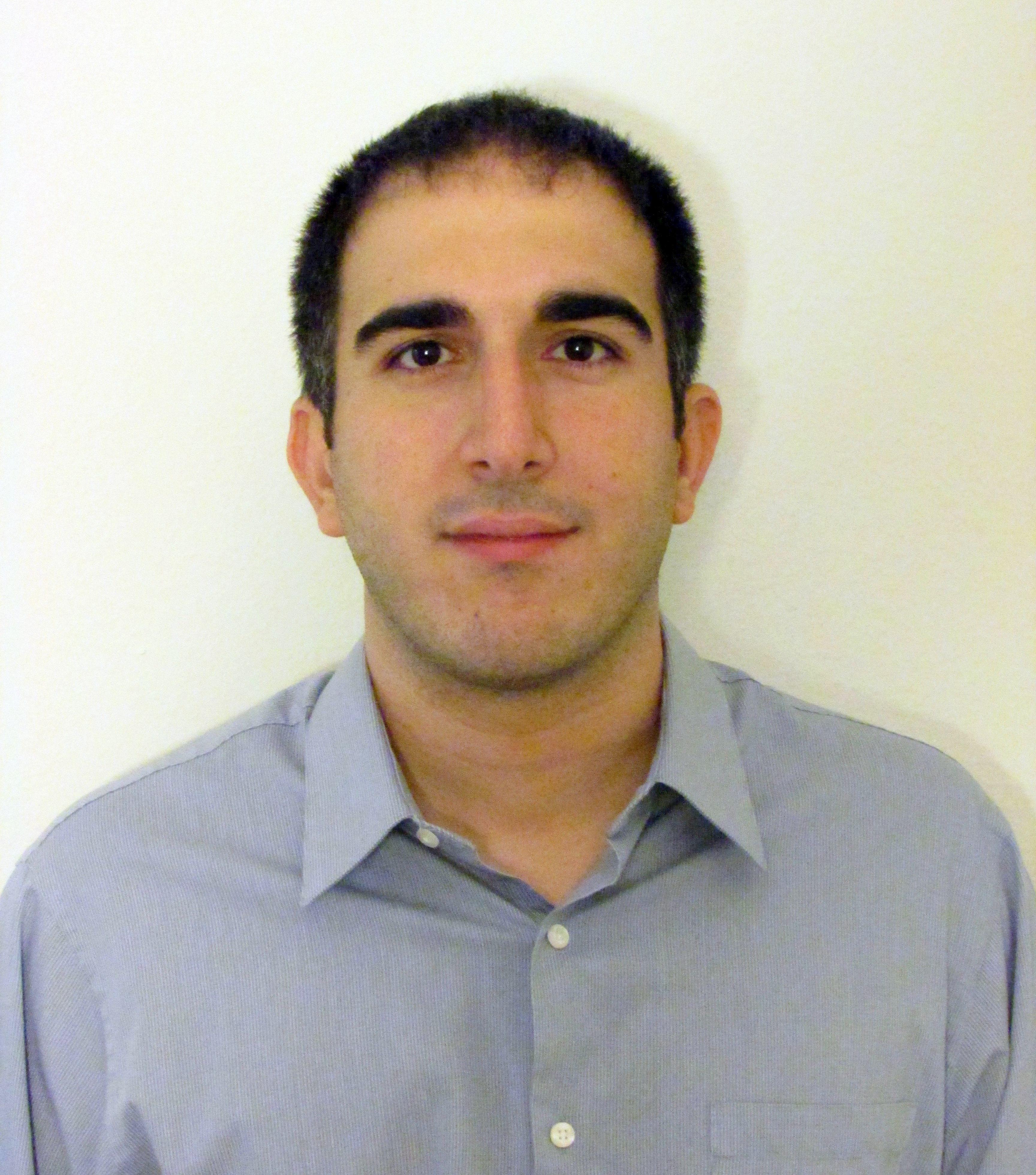}}
\noindent {\bf G\"urkan Solmaz} received his MS and PhD degrees in Computer Science from University of Central Florida (UCF), USA. He received his BS degree in Computer Engineering from Middle East Technical University (METU), Turkey. His research interests include mobility modeling, coverage in mobile wireless sensor networks, and disaster resilience in networks. He is a member of IEEE and ComSoc.

\parpic{\includegraphics[width=1in,clip,keepaspectratio]{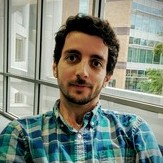}}
\noindent {\bf Rouhollah Rahmatizadeh} received the BS degree in computer engineering from Sharif University of Technology, Tehran, Iran, in 2012 and the MS degree in computer science from the University of Central Florida (UCF), Orlando, in 2014. He is currently pursuing the Ph.D. degree in computer science at UCF. His research interests include machine learning, robotics, and wireless sensor networks.

\parpic{\includegraphics[width=1in,clip,keepaspectratio]{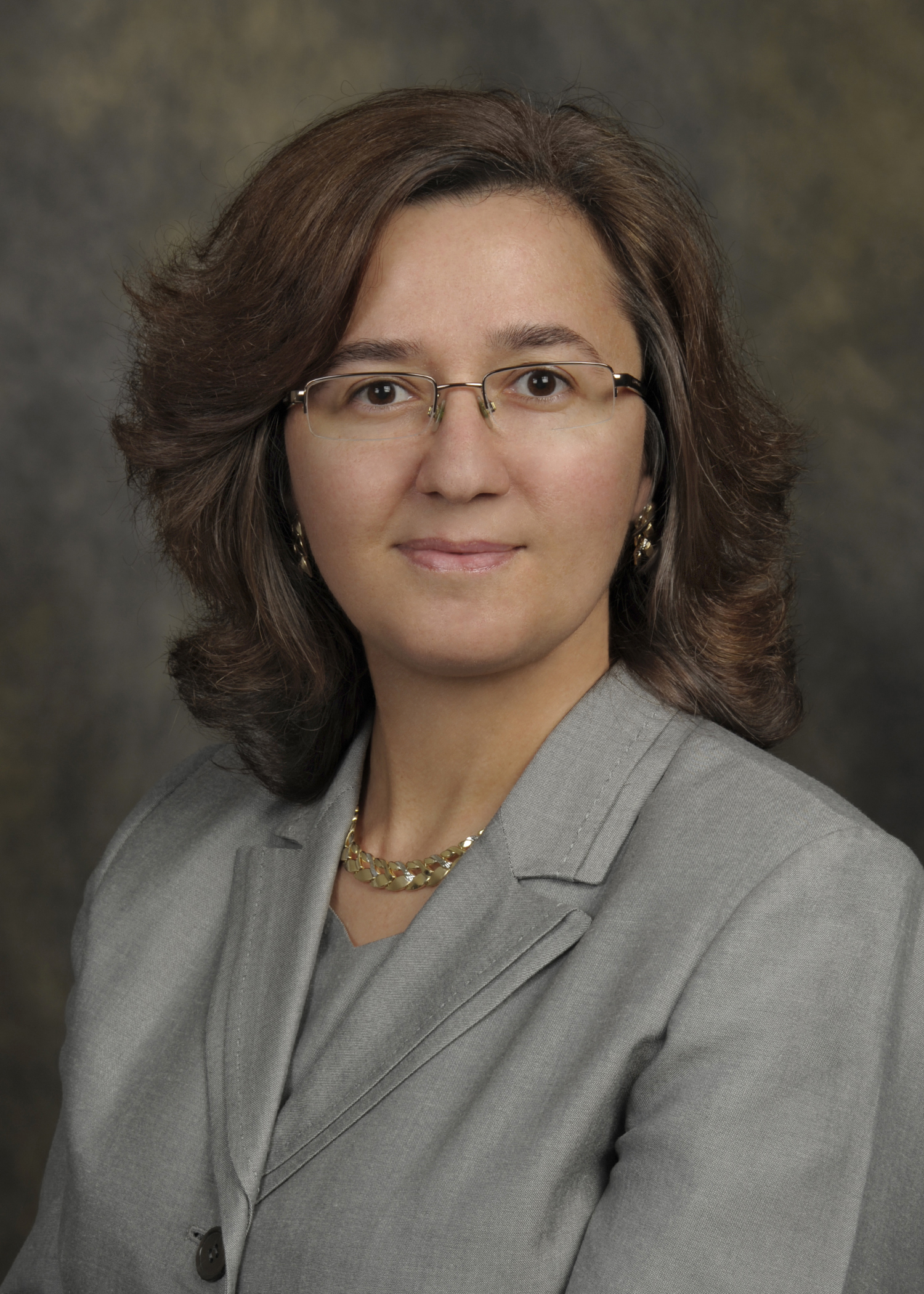}}
\noindent {\bf Damla Turgut} is an Associate Professor at the Department of Electrical Engineering and Computer Science of University of Central Florida. She received her BS, MS, and PhD degrees from the Computer Science and Engineering Department of University of Texas at Arlington. Her research interests include wireless ad hoc, sensor, underwater and vehicular networks, as well as considerations of privacy in the Internet of Things. She is also interested in applying big data techniques for improving STEM education for women and minorities. Her recent honors and awards include being selected as an iSTEM Faculty Fellow for 2014-2015 and being featured in the UCF Women Making History series in March 2015. She was co-recipient of the Best Paper Award at the IEEE ICC 2013. Dr. Turgut serves as a member of the editorial board and of the technical program committee of ACM and IEEE journals and international conferences. She is a member of IEEE, ACM, and the Upsilon Pi Epsilon honorary society.

\parpic{\includegraphics[width=1in,clip,keepaspectratio]{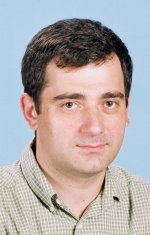}}
\noindent {\bf Ladislau~B{\"o}l{\"o}ni} is an Associate Professor at the Department of Computer Science of University of Central Florida (with a secondary joint appointment in the Dept. of Electrical and Computer Engineering). He received a PhD degree from the Computer Sciences Department of Purdue University in May 2000, an MSc degree from the Computer Sciences department of Purdue University in 1999 and BSc. Computer Engineering with Honors from the Technical University of Cluj-Napoca , Romania in 1993. He received a fellowship from the Computer and Automation Research Institute of the Hungarian Academy of Sciences for the 1994-95 academic year. He is a senior member of IEEE, member of the ACM, AAAI and the Upsilon Pi Epsilon honorary society. His research interests include cognitive science, autonomous agents, grid computing and wireless networking.

\end{document}